\documentclass[10pt,journal,compsoc]{IEEEtran}
\usepackage{ifpdf}
\usepackage[nocompress]{cite}
\usepackage{url}
\usepackage{xcolor}
\usepackage{colortbl}
\usepackage{array}
\usepackage{float}
\usepackage{graphicx}
\usepackage{amssymb}
\usepackage{pifont}
\usepackage{hyperref}
\usepackage{array}
\usepackage{varwidth}
\usepackage{todonotes}
\usepackage{stfloats}
\usepackage{changes}
\hypersetup{linkcolor=blue; citecolor=green;}
\newcommand{\cmark}{\ding{51}}%
\newcommand{\xmark}{\ding{55}}%

\begin{document}

\title{Self-supervised Visual Feature Learning with Deep Neural Networks: A Survey}

\author{Longlong~Jing and Yingli~Tian$^{*}$,~\IEEEmembership{Fellow,~IEEE}      
\IEEEcompsocitemizethanks{\IEEEcompsocthanksitem L. Jing is with the Department of Computer Science, The Graduate Center, The City University of New York, NY, 10016. E-mail: ljing@gradcenter.cuny.edu
\protect\\
\IEEEcompsocthanksitem Y. Tian is with the Department of Electrical Engineering, The City College, and the Department of Computer Science, the Graduate Center, the City University of New York, NY, 10031. E-mail: ytian@ccny.cuny.edu \protect\\
$^*$Corresponding author \protect\\
}
\thanks{This material is based upon work supported by the National Science Foundation under award number IIS-1400802.}}

\markboth{}%
{Shell \MakeLowercase{\textit{\textit{et al.}}}: Bare Demo of IEEEtran.cls for Computer Society Journals}
%


\IEEEtitleabstractindextext{
\begin{abstract}

Large-scale labeled data are generally required to train deep neural networks in order to obtain better performance in visual feature learning from images or videos for computer vision applications. To avoid extensive cost of collecting and annotating large-scale datasets, as a subset of unsupervised learning methods, self-supervised learning methods are proposed to learn general image and video features from large-scale unlabeled data without using any human-annotated labels. This paper provides an extensive review of deep learning-based self-supervised general visual feature learning methods from images or videos. First, the motivation, general pipeline, and terminologies of this field are described. Then the common deep neural network architectures that used for self-supervised learning are summarized. Next, the schema and evaluation metrics of self-supervised learning methods are reviewed followed by the commonly used image and video datasets and the existing self-supervised visual feature learning methods. Finally, quantitative performance comparisons of the reviewed methods on benchmark datasets are summarized and discussed for both image and video feature learning. At last, this paper is concluded and lists a set of promising future directions for self-supervised visual feature learning. 

\end{abstract}

}

\maketitle

\IEEEdisplaynontitleabstractindextext


\IEEEpeerreviewmaketitle

\IEEEraisesectionheading{\section{Introduction}\label{sec:introduction}}


\subsection{Motivation}

\IEEEPARstart{D}{ue} to the powerful ability to learn different levels of general visual features, deep neural networks have been used as the basic structure to many computer vision applications such as object detection \cite{rcnn, fastrcnn, fasterrcnn}, semantic segmentation \cite{FCN, DeepLab, PSPNet}, image captioning \cite{ShowTell}, etc. The models trained from large-scale image datasets like ImageNet are widely used as the pre-trained models and fine-tuned for other tasks for two main reasons: (1) the parameters learned from large-scale diverse datasets provide a good starting point, therefore, networks training on other tasks can converge faster, (2) the network trained on large-scale datasets already learned the hierarchy features which can help to reduce over-fitting problem during the training of other tasks, especially when datasets of other tasks are small or training labels are scarce. 

The performance of deep convolutional neural networks (ConvNets) greatly depends on their capability and the amount of training data. Different kinds of network architectures were developed to increase the capacity of network models, and larger and larger datasets were collected these days. Various networks including AlexNet \cite{AlexNet}, VGG \cite{VGG}, GoogLeNet \cite{GoogLeNet}, ResNet \cite{ResNet}, and DenseNet \cite{DenseNet} and large scale datasets such as ImageNet \cite{ImageNet}, OpenImage \cite{OpenImage} have been proposed to train very deep ConvNets. With the sophisticated architectures and large-scale datasets, the performance of ConvNets keeps breaking the state-of-the-arts for many computer vision tasks \cite{SRGAN, rcnn, FCN, ShowTell, C3D}.

However, collection and annotation of large-scale datasets are time-consuming and expensive. As one of the most widely used datasets for pre-training very deep 2D convolutional neural networks (2DConvNets), ImageNet \cite{ImageNet} contains about $1.3$ million labeled images covering $1,000$ classes while each image is labeled by human workers with one class label. Compared to image datasets, collection and annotation of video datasets are more expensive due to the temporal dimension. The Kinetics dataset \cite{Kinetics}, which is mainly used to train ConvNets for video human action recognition, consists of $500,000$ videos belonging to $600$ categories and each video lasts around $10$ seconds. It took many Amazon Turk workers a lot of time to collect and annotate a dataset at such a large scale.

\begin{figure}[!ht]
\begin{center}
\includegraphics[width=0.4\textwidth]{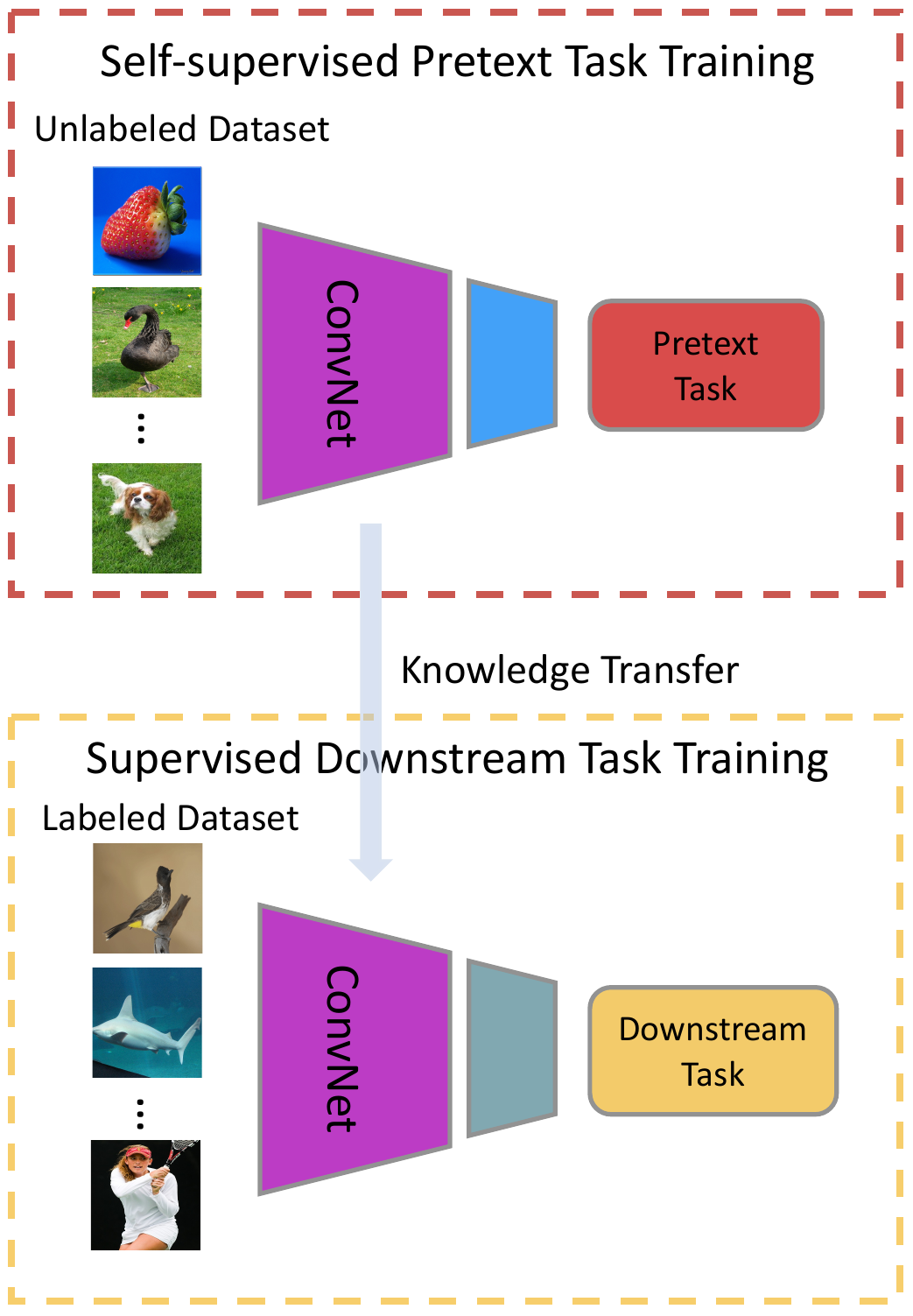}
\end{center}
\caption{The general pipeline of self-supervised learning. The visual feature is learned through the process of training ConvNets to solve a pre-defined pretext task. After self-supervised pretext task training finished, the learned parameters serve as a pre-trained model and are transferred to other downstream computer vision tasks by fine-tuning. The performance on these downstream tasks is used to evaluate the quality of the learned features. During the knowledge transfer for downstream tasks, the general features from only the first several layers are unusually transferred to downstream tasks.}
\label{fig:pipeline}
\end{figure}

To avoid time-consuming and expensive data annotations, many self-supervised methods were proposed to learn visual features from large-scale unlabeled images or videos without using any human annotations. To learn visual features from unlabeled data, a popular solution is to propose various pretext tasks for networks to solve, while the networks can be trained by learning objective functions of the pretext tasks and the features are learned through this process. Various pretext tasks have been proposed for self-supervised learning including colorizing grayscale images \cite{colorfulcolorization}, image inpainting \cite{contextencoder}, image jigsaw puzzle \cite{Jigsaw}, etc. The pretext tasks share two common properties: (1) visual features of images or videos need to be captured by ConvNets to solve the pretext tasks, (2) pseudo labels for the pretext task can be automatically generated based on the attributes of images or videos.

The general pipeline of self-supervised learning is shown in Fig.~\ref{fig:pipeline}. During the self-supervised training phase, a pre-defined pretext task is designed for ConvNets to solve, and the pseudo labels for the pretext task are automatically generated based on some attributes of data. Then the ConvNet is trained to learn object functions of the pretext task. After the self-supervised training finished, the learned visual features can be further transferred to downstream tasks (especially when only relatively small data available) as pre-trained models to improve performance and overcome over-fitting.  Generally, shallow layers capture general low-level features like edges, corners, and textures while deeper layers capture task related high-level features. Therefore, visual features from only the first several layers are transferred during the supervised downstream task training phase.

\subsection{Term Definition}

To make this survey easy to read, we first define the terms used in the remaining sections.

\begin{itemize}

\item \textbf{Human-annotated label: } Human-annotated labels refer to labels of data that are manually annotated by human workers. 

\item \textbf{Pseudo label: } Pseudo labels are automatically generated labels based on data attributes for pretext tasks.

\item \textbf{Pretext Task: } Pretext tasks are pre-designed tasks for networks to solve, and visual features are learned by learning objective functions of pretext tasks.

\item \textbf{Downstream Task: } Downstream tasks are computer vision applications that are used to evaluate the quality of features learned by self-supervised learning. These applications can greatly benefit from the pre-trained models when training data are scarce. In general, human-annotated labels are needed to solve the downstream tasks. However, in some applications, the downstream task can be the same as the pretext task without using any human-annotated labels.

\item \textbf{Supervised Learning:} Supervised learning indicates learning methods using data with fine-grained human-annotated labels to train networks. 

\item \textbf{Semi-supervised Learning: } Semi-supervised learning refers to learning methods using a small amount of labeled data in conjunction with a large amount of unlabeled data.

\item \textbf{Weakly-supervised Learning: } Weakly supervised learning refers to learning methods to learn with coarse-grained labels or inaccurate labels. The cost of obtaining weak supervision labels is generally much cheaper than fine-grained labels for supervised methods.

\item \textbf{Unsupervised Learning:} Unsupervised learning refers to learning methods without using any human-annotated labels.

\item \textbf{Self-supervised Learning:} Self-supervised learning is a subset of unsupervised learning methods. Self-supervised learning refers to learning methods in which ConvNets are explicitly trained with automatically generated labels. This review only focuses on self-supervised learning methods for visual feature learning with ConvNets in which the features can be transferred to multiple different computer vision tasks.

\end{itemize}

Since no human annotations are needed to generate pseudo labels during self-supervised training, very large-scale datasets can be used for self-supervised training. Trained with these pseudo labels, self-supervised methods achieved promising results and the gap with supervised methods in performance on downstream tasks becomes smaller. This paper provides a comprehensive survey of deep ConvNets-based self-supervised visual feature learning methods. 
The key contributions of this paper are as follows:

\begin{itemize}
 
\item To the best of our knowledge, this is the first comprehensive survey about self-supervised visual feature learning with deep ConvNets which will be helpful for researchers in this field.

\item An in-depth review of recently developed self-supervised learning methods and datasets.

\item Quantitative performance analysis and comparison of the existing methods are provided. 

\item A set of possible future directions for self-supervised learning is pointed out.  

\end{itemize}

\section{Formulation of Different Learning Schemas}

Based on the training labels, visual feature learning methods can be grouped into the following four categories: supervised, semi-supervised, weakly supervised, and unsupervised. In this section, the four types of learning methods are compared and key terminologies are defined. 

\subsection{Supervised Learning Formulation}

For supervised learning, given a dataset X, for each data $X_i$ in X, there is a corresponding \textbf{human-annotated label} $Y_i$. For a set of $N$ labeled training data $D = \{X_i\}_{i=0}^{N}$, the training loss function is defined as:
\begin{equation} \label{eq:supervsied-loss}
loss(D) = \min_{\theta} \frac{1}{N}\sum_{i=1}^{N} loss(X_i,Y_i).
\end{equation}

Trained with accurate human-annotated labels, the supervised learning methods obtained break-through results on different computer vision applications \cite{AlexNet, FCN, rcnn, C3D}. However, data collection and annotation usually are expensive and may require special skills. Therefore, semi-supervised, weakly supervised, and unsupervised learning methods were proposed to reduce the cost.

\subsection{Semi-Supervised Learning Formulation}

For semi-supervised visual feature learning, given a small labeled dataset $X$ and a large unlabeled dataset $Z$, for each data $X_i$ in X, there is a corresponding \textbf{human-annotated label} $Y_i$. For a set of $N$ labeled training data $D_1 = \{X_i\}_{i=0}^{N}$ and $M$ unlabeled training data $D_2 = \{Z_i\}_{i=0}^{M}$, the training loss function is defined as:
\begin{equation} 
\small
\label{eq:semi-supervsied-loss}
loss(D_1, D_2) = \min_{\theta} \frac{1}{N}\sum_{i=1}^{N}loss(X_i,Y_i) + \frac{1}{M}\sum_{i=1}^{M}loss(Z_i,R(Z_i,X)), 
\end{equation} 
where the $R(Z_i,X)$ is a task-specific function to represent the relation between each unlabeled training data $Z_i$ with the labeled dataset $X$. 
 
\subsection{Weakly Supervised Learning Formulation}

For weakly supervised visual feature learning, given a dataset X, for each data $X_i$ in X, there is a corresponding \textbf{coarse-grained label} $C_i$. For a set of $N$ training data $D = \{X_i\}_{i=0}^{N}$, the training loss function is defined as:
\begin{equation} \label{eq:weakly-supervsied-loss}
loss(D) = \min_{\theta} \frac{1}{N}\sum_{i=1}^{N} loss(X_i,C_i).
\end{equation}

Since the cost of weak supervision is much lower than the fine-grained label for supervised methods, large-scale datasets are relatively easier to obtain. Recently, several papers proposed to learn image features from web collected images using hashtags as category labels \cite{weaklylimitation, WebVision}, and obtained very good performance \cite{weaklylimitation}. 

\subsection{Unsupervised Learning Formulation}

Unsupervised learning refers to learning methods that do not need any human-annotated labels. This type of methods including fully unsupervised learning methods in which the methods do not need any labels at all, as well as self-supervised learning methods in which networks are explicitly trained with automatically generated pseudo labels without involving any human annotation.

\subsubsection{Self-supervised Learning}
Recently, many self-supervised learning methods for visual feature learning have been developed without using any human-annotated labels \cite{crosspixel, crossmodel, AVTS, AudioVisual, CubicPuzzles, 3DRotNet, O3N, crossdomain, transitive, multitasklearning, ImproveContext, boosting, ImproveContext, RL}. Some papers refer to this type of learning methods as unsupervised learning \cite{RotNet, Self-LSTM, wang2015unsupervised, sortsequence, shuffleandlearn, contextprediction, splitbrain, graphconstraint, deepcluster, spatialcontrast, predictnoise, unsupervisededges, poseaction}. 
Compared to supervised learning methods which require a data pair $X_i$ and $Y_i$ while $Y_i$ is annotated by human labors, self-supervised learning also trained with data $X_i$ along with its \textbf{pseudo label} $P_i$ while $P_i$ is automatically generated for a pre-defined pretext task without involving any human annotation. The pseudo label $P_i$ can be generated by using attributes of images or videos such as the context of images \cite{Jigsaw, RotNet, colorfulcolorization, contextencoder}, or by traditional hand-designed methods \cite{FisherVector, NCL, VideoPCA}. 

Given a set of $N$ training data $D = \{P_i\}_{i=0}^{N}$, the training loss function is defined as:
\begin{equation} \label{eq:selfsupervsiedloss}
loss(D) = \min_{\theta} \frac{1}{N}\sum_{i=1}^{N} loss(X_i,P_i).
\end{equation}

As long as the pseudo labels $P$ are  automatically generated without involving human annotations, then the methods belong to self-supervised learning. Recently, self-supervised learning methods have achieved great progress. This paper focuses on the self-supervised learning methods that mainly designed for visual feature learning, while the features have the ability to be transferred to multiple visual tasks and to perform new tasks by learning from limited labeled data. This paper summarizes these self-supervised feature learning methods from different perspectives including network architectures, commonly used pretext tasks, datasets, and applications, etc.

\section{Common Deep Network Architectures}

No matter the categories of learning methods, they share similar network architectures. This section reviews common architectures for learning both image and video features.

\subsection{Architectures for Learning Image Features}

Various 2DConvNets have been designed for image feature learning. Here, five milestone architectures for image feature learning including AlexNet \cite{AlexNet}, VGG \cite{VGG}, GoogLeNet \cite{GoogLeNet}, ResNet \cite{ResNet}, and DenseNet \cite{DenseNet} are reviewed.

\subsubsection{AlexNet}

\begin{figure}[!ht]
\begin{center}
\includegraphics[width=0.37\textwidth]{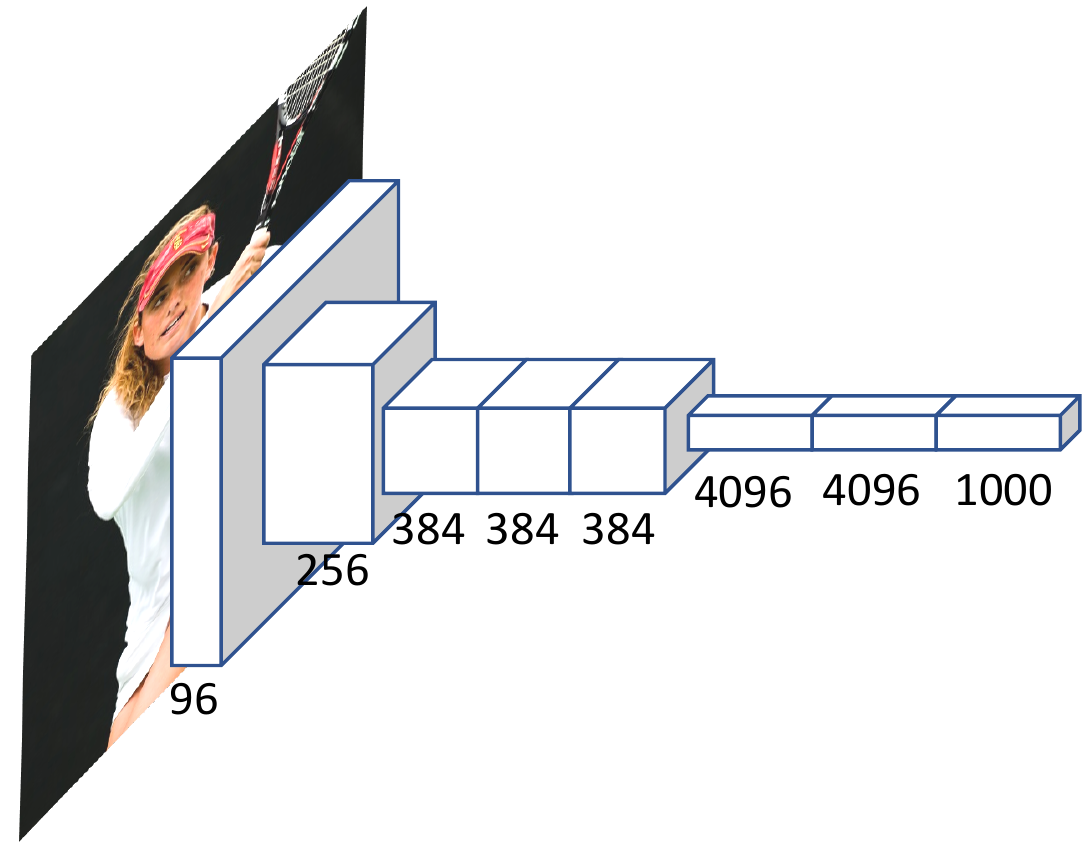}
\end{center}
\caption{The architecture of AlexNet \cite{AlexNet}. The numbers indicate the number of channels of each feature map. Figure is reproduced based on AlexNet \cite{AlexNet}.}
\label{fig:AlexNet}
\end{figure}

AlexNet obtained a big improvement in the performance of image classification on ImageNet dataset compared to the previous state-of-the-art methods \cite{AlexNet}. With the support of powerful GPUs,  AlexNet which has $62.4$ million parameters were trained on ImageNet with $1.3$ million images. As shown in Fig.~\ref{fig:AlexNet}, the architecture of AlexNet has $8$ layers in which $5$ are convolutional layers and $3$ are fully connected layers. The ReLU is applied after each convolutional layers. $94$\% of the network parameters come from the fully connected layers. With this scale of parameters, the network can easily be over-fitting. Therefore, different kinds of techniques are applied to avoid over-fitting problem including data augmentation, dropout, and normalization.

\subsubsection{VGG} 

\begin{figure}[!ht]
\begin{center}
\includegraphics[width=0.5\textwidth]{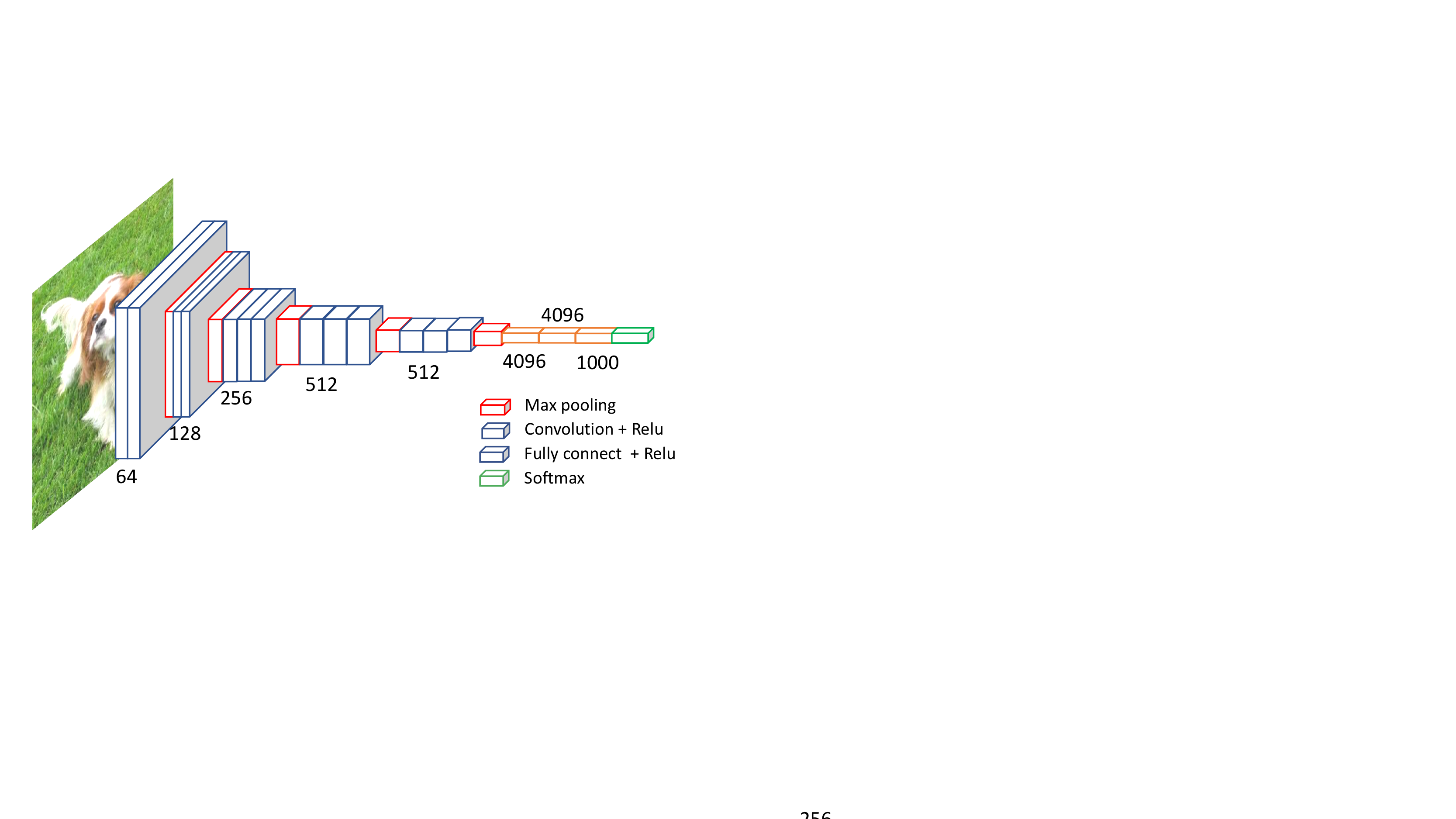}
\end{center}
\caption{The architecture of VGG \cite{VGG}. Figure is reproduced based on VGG \cite{VGG}.}
\label{fig:VGG}
\end{figure}

VGG is proposed by Simonyan and Zisserman and won the first place for ILSVRC 2013 competition \cite{VGG}. Simonyan and Zisserman proposed various depth of networks, while the 16-layer VGG is the most widely used one due to its moderate model size and its superior performance. The architecture of VGG-16 is shown in Fig.~\ref{fig:VGG}. It has $16$ convolutional layers belong to five convolution blocks. The main difference between VGG and AlexNet is that AlexNet has large convolution stride and large kernel size while all the convolution kernels in VGG have same small size ($3\times3$) and small convolution stride ($1\times1$). The large kernel size leads to too many parameters and large model size, while the large convolution stride may cause the network to miss some fine features in the lower layers. The smaller kernel size makes the training of very deep convolution neural network feasible while still reserving the fine-grained information in the network.

\subsubsection{ResNet}

\begin{figure}[!ht]
\begin{center}
\includegraphics[width=0.31\textwidth]{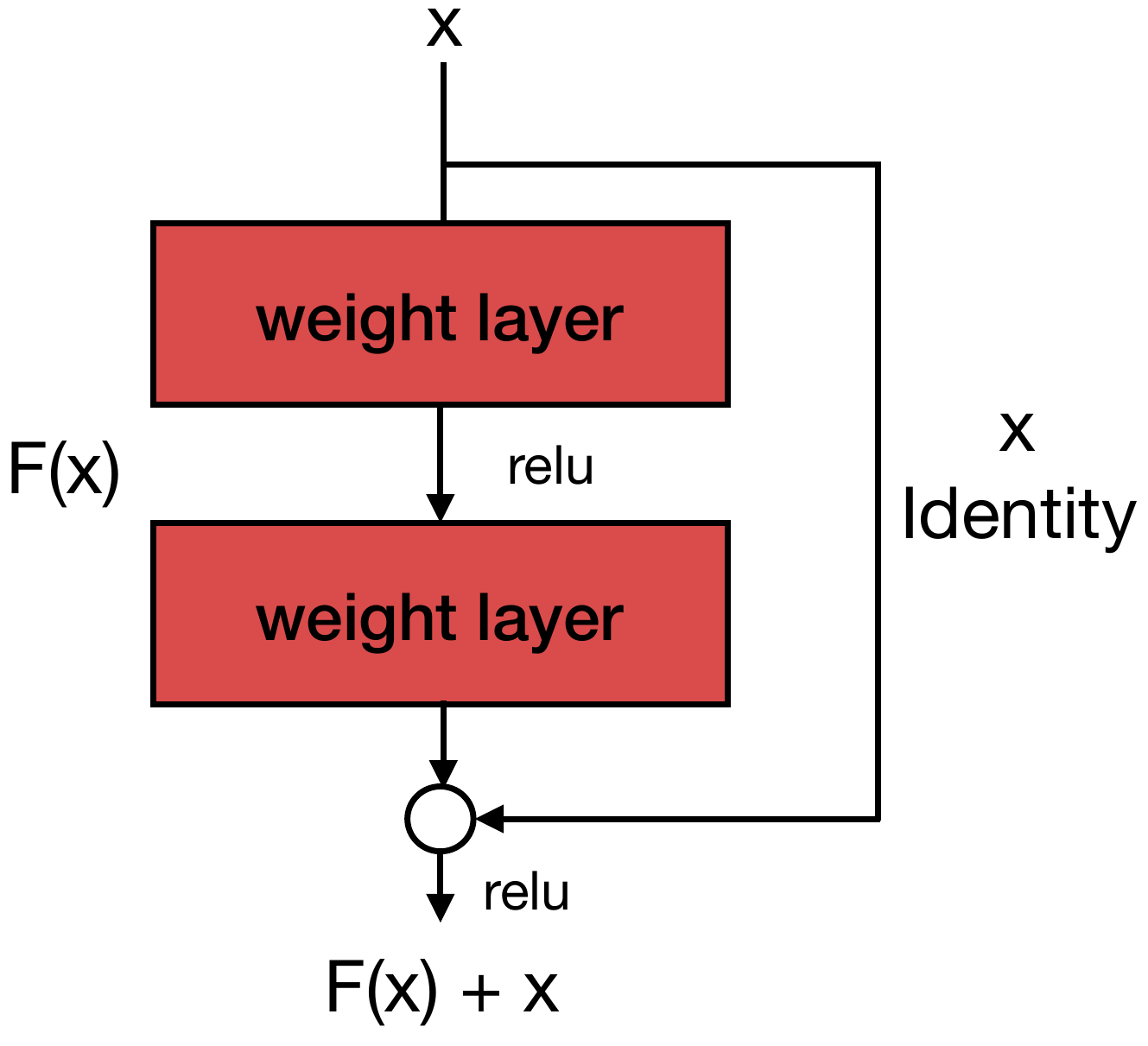}
\end{center}
\caption{The architecture of Residual block \cite{ResNet}. The identity mapping can effectively reduce gradient vanishing and explosion which make the training of very deep network feasible. Figure is reproduced based on ResNet \cite{ResNet}.}
\label{fig:ResNet}
\end{figure}

VGG demonstrated that deeper networks are possible to obtain better performance. However, deeper networks are more difficult to train due to two problems: gradient vanishing and gradient explosion. ResNet is proposed by He \textit{et al.} to use the skip connection in convolution blocks by sending the previous feature map to the next convolution block to overcome the gradient vanishing and gradient explosion \cite{ResNet}. The details of the skip connection are shown in Fig.~\ref{fig:ResNet}. With the skip connection, training of very deep neural networks on GPUs becomes feasible. 

In ResNet \cite{ResNet}, He \textit{et al.} also evaluated networks with different depths for image classification. Due to its smaller model size and superior performance, ResNet is often used as the base network for other computer vision tasks. The convolution blocks with skip connection also widely used as the basic building blocks.

\subsubsection{GoogLeNet}

\begin{figure}[!ht]
\begin{center}
\includegraphics[width=0.5\textwidth]{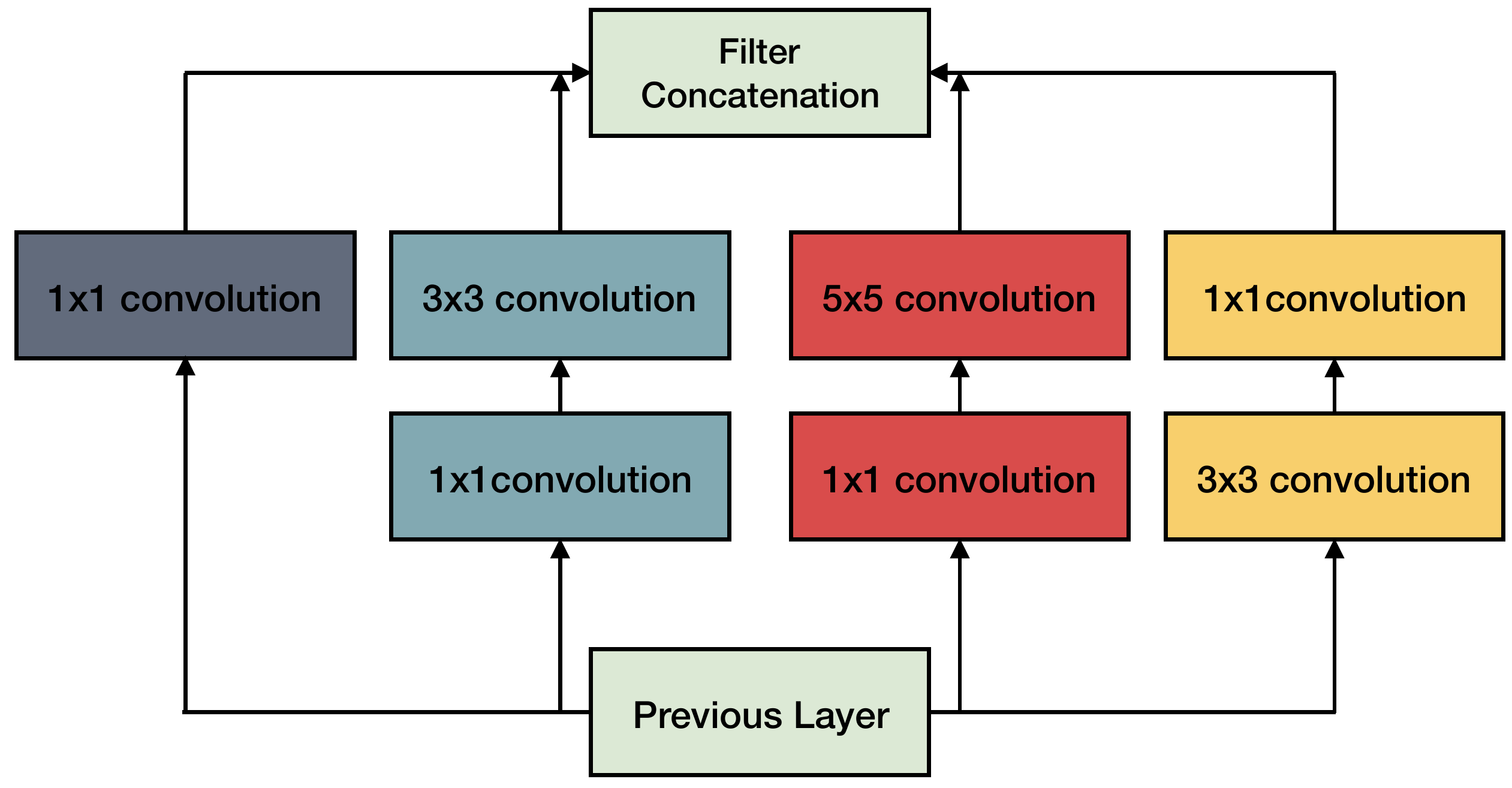}
\end{center}
\caption{The architecture of Inception block \cite{GoogLeNet}. Figure is reproduced based on GoogLeNet \cite{GoogLeNet}.}
\label{fig:GoogLeNet}
\end{figure}

GoogLeNet, a $22$-layer deep network, is proposed by Szegedy \textit{et al.}  which won ILSVRC-2014 challenge with a top-5 test accuracy of $93.3$\% \cite{GoogLeNet}. Compared to previous work that to build a deeper network, Szegedy \textit{et al.} explored to build a wider network in which each layer has multiple parallel convolution layers. The basic block of GoogLeNet is inception block which consists of $4$ parallel convolution layers with different kernel sizes and followed by $1 \times 1$ convolution for dimension reduction purpose. The architecture for the inception block of GoogLeNet is shown in Fig.~\ref{fig:GoogLeNet}. With a carefully crafted design, they increased the depth and width of the network while keeping the computational cost constant.

\subsubsection{DenseNet}

\begin{figure}[!ht]
\begin{center}
\includegraphics[width=0.5\textwidth]{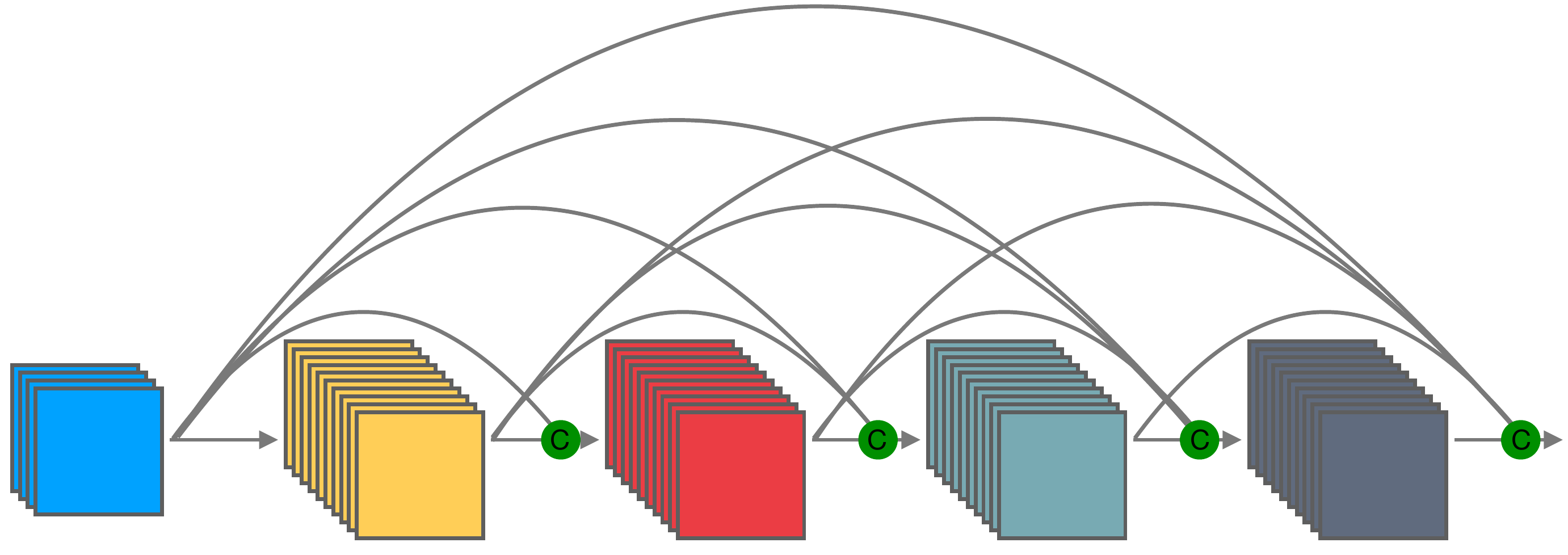}
\end{center}
\caption{The architecture of the Dense Block proposed in DenseNet \cite{DenseNet}. Figure is reproduced based on \cite{DenseNet}.}
\label{fig:densenet}
\end{figure}

Most of the networks including AlexNet, VGG, and ResNet follow a hierarchy architecture. The images are fed to the network and features are extracted by different layers. The shallow layers extract low-level general features, while the deep layers extract high-level task-specific features \cite{understanding}. However, when a network goes deeper, the deeper layers may suffer from memorizing the low-level features needed by the network to accomplish the task.

To alleviate this problem, Huang \textit{et al.} proposed the dense connection to send all the features before a convolution block as the input to the next convolution block in the neural network \cite{DenseNet}. As shown in Fig.~\ref{fig:densenet}, the output features of all the previous convolution blocks serve as the input to the current block. In this way, the shallower blocks focus on the low-level general features while the deeper blocks can focus on the high-level task-specific features.

\subsection{Architectures for Learning Video Features}

To extract both spatial and temporal information from videos, several architectures have been designed for video feature learning including 2DConvNet-based methods \cite{Two-Stream}, 3DConvNet-based methods \cite{C3D}, and LSTM-based methods \cite{LRCN}. The 2DConvNet-based methods apply 2DConvNet on every single frame and the image features of multiple frames are fused as video features. The 3DConvNet-based methods employ 3D convolution operation to simultaneously extract both spatial and temporal features from multiple frames. The LSTM-based methods employ LSTM to model long term dynamics within a video. This section briefly summarizes these three types of architectures of video feature learning.

\subsubsection{Two-Stream Network}

\begin{figure}[!ht]
\begin{center}
\includegraphics[width=0.45\textwidth]{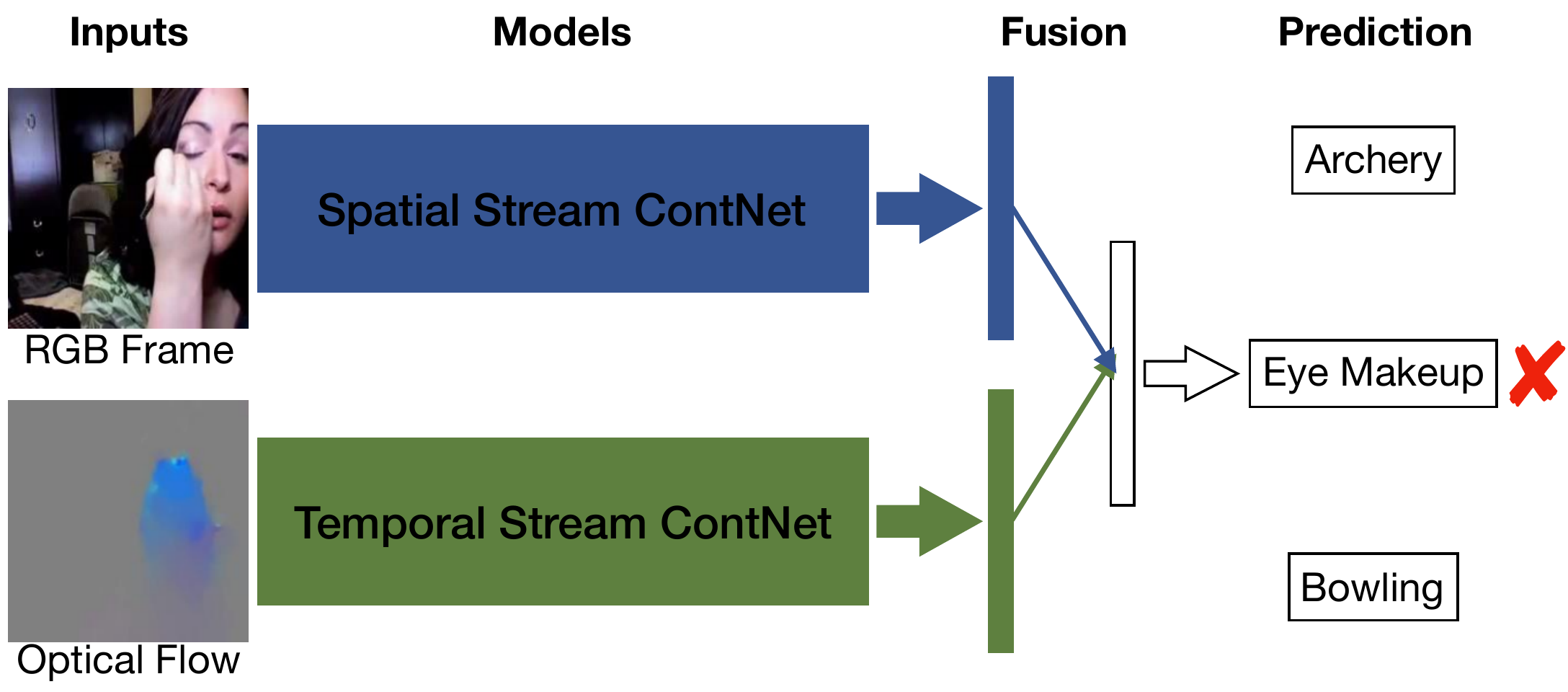}
\end{center}
\caption{The general architecture of the two-stream network which including one spatial stream and one temporal stream. Figure is reproduced based on \cite{Two-Stream}.}
\label{fig:two-stream}
\end{figure}

Videos generally are composed of various numbers of frames. To recognize actions in a video, networks are required to capture appearance features as well as temporal dynamics from frame sequences. As shown in Fig.~\ref{fig:two-stream}, a two-stream 2DConvNet-based network is proposed by Simonyan and Zisserman for human action recognition, while using a 2DConvNet to capture spatial features from RGB stream and another 2DConvNet to capture temporal features from optical flow stream \cite{Two-Stream}. Optical flow encodes boundary of moving objects, therefore, the temporal stream ConvNet is relatively easier to capture the motion information within the frames. 

Experiments showed that the fusion of the two streams can significantly improve action recognition accuracy. Later, this work has been extended to multi-stream network \cite{two-stream-one, two-stream-two, two-stream-three, two-stream-four, two-stream-five} to fuse features from different types of inputs such as dynamic images \cite{dynamicimage} and difference of frames \cite{TSN}.

\subsubsection{Spatiotemporal Convolutional Neural Network}

3D convolution operation was first proposed in 3DNet \cite{3D} for human action recognition. Compared to 2DConvNets which individually extract the spatial information of each frame and then fuse them together as video features, 3DConvNets are able to simultaneously extract both spatial and temporal features from multiple frames. 

C3D \cite{C3D} is a VGG-like 11-layer 3DConvNet designed for human action recognition. The network contains $8$ convolutional layers, and $3$ fully connected layers. All the kernels have the size of $3 \times 3 \times 3$, the convolution stride is fixed to $1$ pixel. Due to its powerful ability of simultaneously extracting both spatial and temporal features from multiple frames, the network achieved state-of-the-art on several video analysis tasks including human action recognition \cite{UCF101}, action similarity labeling \cite{C3D-ASLAN}, scene classification \cite{C3D-YUPENN}, and object recognition in videos \cite{C3D-Object}. 

The input of C3D is $16$ consecutive RGB frames where the appearance and temporal cues from  16-frame clips are extracted. However, the paper of long-term temporal convolutions (LTC) \cite{LTC} argues that, for the long-lasting actions, 16 frames are insufficient to represent whole actions which last longer. Therefore, larger numbers of frames are employed to train 3DConvNets and achieved better performance than C3D \cite{LTC, VideoYOLO}. 

With the success of applying 3D convolution on video analysis tasks, various 3DConvNet architectures have been proposed \cite{I3D, 3DResNet, P3D}. Hara \textit{et al.} proposed 3DResNet by replacing all the 2D convolution layers in ResNet with 3D convolution layers and showed comparable performance with the state-of-the-art performance on action recognition task on several datasets \cite{3DResNet}. 

\subsubsection{Recurrent Neural Network}

\begin{figure}[!ht]
\begin{center}
\includegraphics[width=0.38\textwidth]{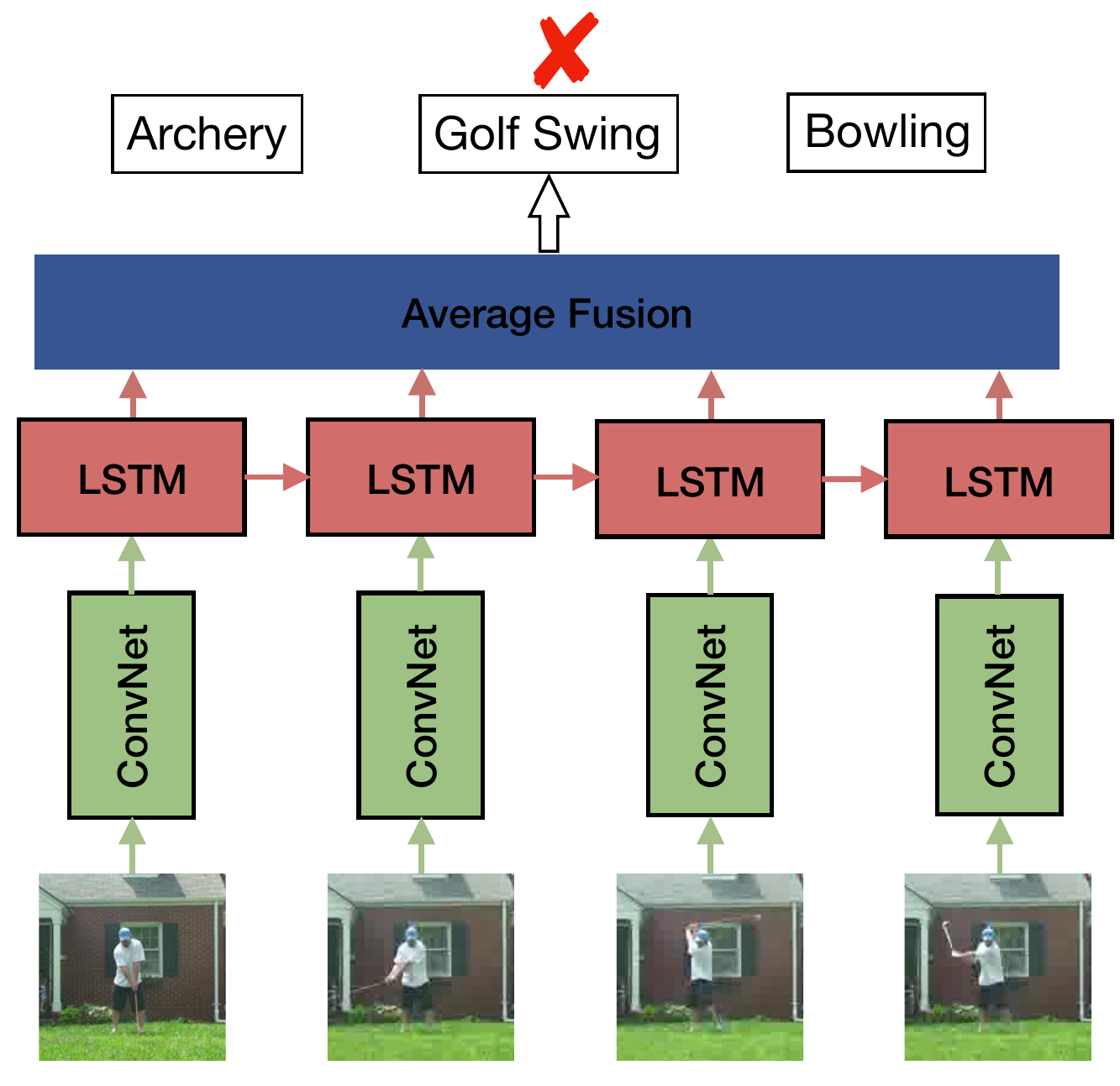}
\end{center}
\caption{The architecture of long-term recurrent convolutional
networks (LRCN) \cite{LRCN}. LSTM is employed to model the long term temporal information within a frame sequence. Figure is reproduced based on \cite{LRCN}.}
\label{fig:LRCN}
\end{figure}

Due to the ability to model the temporal dynamics within a sequence, recurrent neural networks (RNN) are often applied to videos as ordered frame sequences. Compared to standard RNN \cite{RNN}, long short term memory (LSTM) uses memory cells to store, modify, and access internal states, to better model the long-term temporal relationships within video frames \cite{LSTM}.

Based on the advantage of the LSTM, Donahue \textit{et al.} proposed long-term recurrent convolutional networks (LRCN) for human action recognition \cite{LRCN}. The framework of the LRCN is shown in Fig.~\ref{fig:LRCN}. The LSTM is sequentially applied to the features extracted by ConvNets to model the temporal dynamics in the frame sequence. With the LSTM to model a video as frame sequences, this model is able to explicitly model the long-term temporal dynamics within a video. Later on, this model is extended to a deeper LSTM for action recognition \cite{Beyond, VideoLSTM}, video captioning \cite{V2T}, and gesture recognition tasks \cite{gesturesLSTM}.

\subsection{Summary of ConvNet Architectures}

Deep ConvNets have demonstrated great potential in various computer vision tasks. And the visualization of the image and video features has shown that these networks truly learned meaningful features that required by the corresponding tasks \cite{understanding, NetDissection, GANDissection, visualizeunderstandcnn}. However, one common drawback is that these networks can be easily over-fit when training data are scarce since there are over millions of parameters in each network. 

Take 3DResNet for an example, the performance of an $18$-layer 3DResNet on UCF101 action recognition dataset \cite{UCF101} is $42$\% when trained from scratch. However, with a supervised pre-trained model on the large-scale Kinetics dataset ($500,000$ videos of $600$ classes) with human-annotated class labels and then fine-tuned on UCF101 dataset, the performance can increase to $84$\%. Pre-trained models on large-scale datasets can speed up the training process and improve the performance on relatively small datasets. However, the cost of collecting and annotating large-scale datasets is very expensive and time-consuming.

In order to obtain pre-trained models from large-scale datasets without expensive human annotations, many self-supervised learning methods were proposed to learn image and video features from pre-designed pretext tasks. The next section describes the general pipeline of the self-supervised image and video feature learning.

\section{Commonly used Pretext and Downstream Tasks}

\begin{figure}[!ht]
\begin{center}
\includegraphics[width=0.5\textwidth]{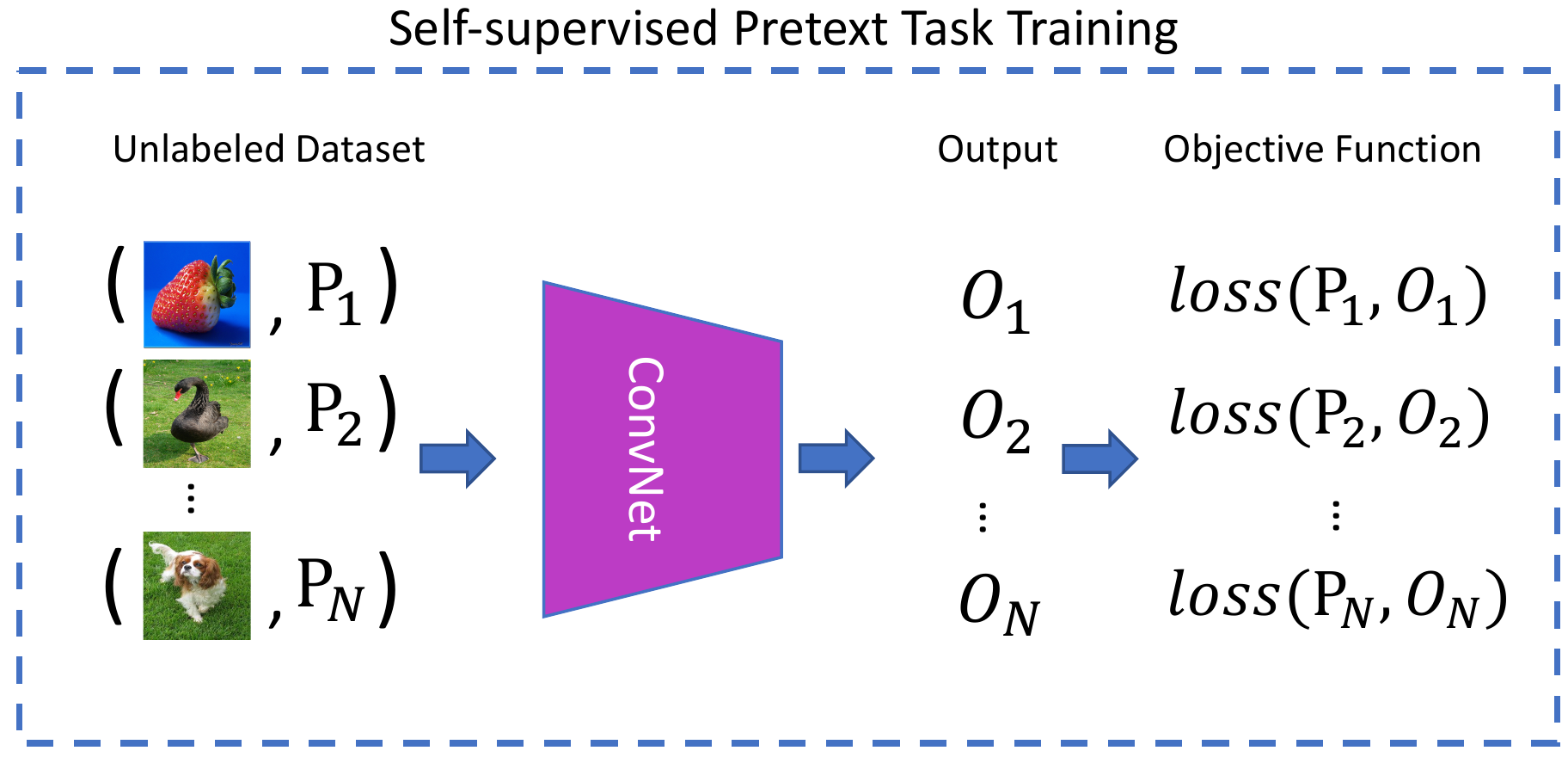}
\end{center}
\caption{Self-supervised visual feature learning schema. The ConvNet is trained by minimizing errors between pseudo labels $P$ and predictions $O$ of the ConvNet. Since the pseudo labels are automatically generated, no human annotations are involved during the whole process.}
\label{fig:selfsupervisepipeline}
\end{figure}

Most existing self-supervised learning approaches follow the schema shown in Fig~\ref{fig:selfsupervisepipeline}. Generally, a pretext task is defined for ConvNets to solve and visual features can be learned through the process of accomplishing this pretext task. The pseudo labels $P$ for pretext task can be automatically generated without human annotations. ConvNet is optimized by minimizing the error between the prediction of ConvNet $O$ and the pseudo labels $P$. After the training on the pretext task is finished, ConvNet models that can capture visual features for images or videos are obtained. 

\subsection{Learning Visual Features from Pretext Tasks}

To relieve the burden of large-scale dataset annotation, a pretext task is generally designed for networks to solve while pseudo labels for the pretext task are automatically generated based on data attributes. Many pretext tasks have been designed and applied for self-supervised learning such as foreground object segmentation \cite{watchingmove}, image inpainting \cite{contextencoder}, clustering \cite{deepcluster}, image colorization \cite{colorproxy}, temporal order verification \cite{shuffleandlearn}, visual audio correspondence verification \cite{AVTS}, and so on. Effective pretext tasks ensure that semantic features are learned through the process of accomplishing the pretext tasks. 

Take image colorization as an example, image colorization is a task to colorize gray-scale images into colorful images. To generate realistic colorful images, networks are required to learn the structure and context information of images. In this pretext task, the data $X$ is the gray-scale images which can be generated by performing a linear transformation in RGB images, while the pseudo label $P$ is the RGB image itself. The training pair $X_i$ and $P_i$ can be generated in real time with negligible cost. Self-Supervised learning with other pretext tasks follow a similar pipeline.

\subsection{Commonly Used Pretext Tasks}

\begin{figure*}[!ht]
\begin{center}
\includegraphics[width=\textwidth]{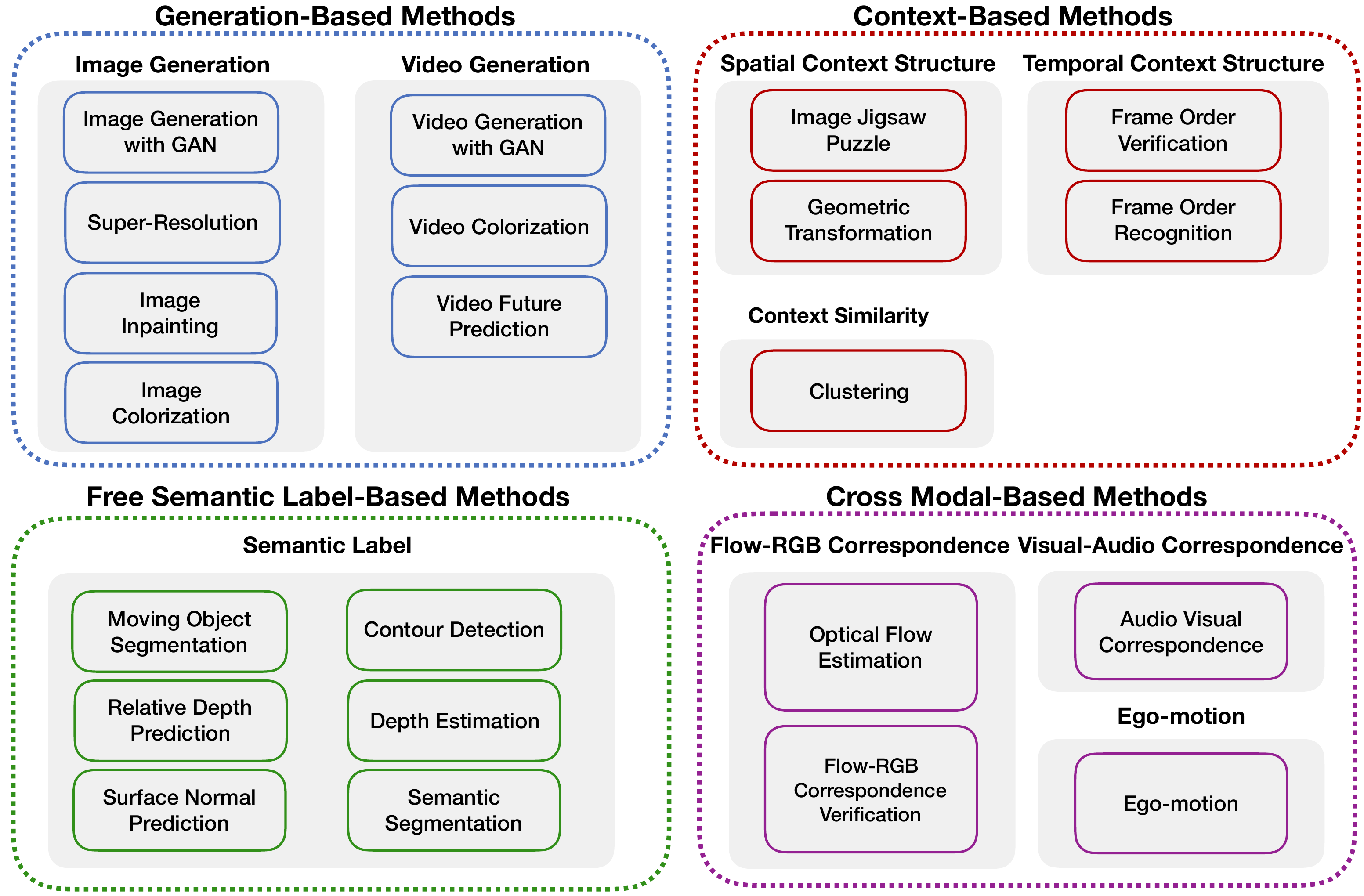}
\caption{Categories of pretext tasks for self-supervised visual feature learning: generation-based, context-based, free semantic label-based, and cross modal-based. }

\label{fig:graph}
\end{center}
\end{figure*}
 
According to the data attributes used to design pretext tasks, as shown in Fig.~\ref{fig:graph}, we summarize the pretext tasks into four categories: generation-based, context-based, free semantic label-based, and cross modal-based.

\textbf{Generation-based Methods:} This type of methods learn visual features by solving pretext tasks that involve image or video generation.

\begin{itemize}
  
  \item{\textbf{Image Generation:} Visual features are learned through the process of image generation tasks. This type of methods includes image colorization \cite{colorfulcolorization}, image super resolution \cite{SRGAN}, image inpainting \cite{contextencoder}, image generation with Generative Adversarial Networks (GANs)  \cite{GAN, CycleGAN}.}

  \item{\textbf{Video Generation:} Visual features are learned through the process of video generation tasks. This type of methods includes video generation with GANs \cite{VideoGAN, MocoGAN} and video prediction \cite{Self-LSTM}.}
  
\end{itemize}

\textbf{Context-based pretext tasks:} The design of context-based pretext tasks mainly employ the context features of images or videos such as context similarity, spatial structure, temporal structure, etc. 

\begin{itemize}
  \item{\textbf{Context Similarity:} Pretext tasks are designed based on the context similarity between image patches. This type of methods includes image clustering-based methods \cite{boosting, deepcluster}, and graph constraint-based methods \cite{graphconstraint}.}

  \item{\textbf{Spatial Context Structure:} Pretext tasks are used to train ConvNets are based on the spatial relations among image patches. This type of methods includes image jigsaw puzzle \cite{Jigsaw, VideoJigsaw, ArbitraryJigsaw, DamagedJigsaw}, context prediction \cite{contextprediction}, and geometric transformation recognition \cite{RotNet, 3DRotNet}, etc. } 

  \item{\textbf{Temporal Context Structure:} The temporal order from videos is used as supervision signal. The ConvNet is trained to verify whether the input frame sequence in correct order \cite{shuffleandlearn, arrowoftime} or to recognize the order of the frame sequence \cite{sortsequence}.} 
\end{itemize}

\textbf{Free Semantic Label-based Methods:} This type of pretext tasks train networks with automatically generated semantic labels. The labels are generated by traditional hard-code algorithms \cite{NCL, VideoPCA} or by game engines \cite{crossdomain}. The pretext tasks include moving object segmentation \cite{forground_seg, watchingmove}, contour detection \cite{unsupervisededges, crossdomain}, relative depth prediction \cite{relativedepth}, and etc.

\textbf{Cross Modal-based Methods:} This type of pretext tasks train ConvNets to verify whether two different channels of input data are corresponding to each other. This type of methods includes Visual-Audio Correspondence Verification \cite{looklistenlearn, AVTS}, RGB-Flow Correspondence Verification \cite{crossmodel}, and egomotion \cite{learn2seebymove, TiedEgoMotion}.

\subsection{Commonly Used Downstream Tasks for Evaluation}

To evaluate the quality of the learned image or video features by self-supervised methods, the learned parameters by self-supervised learning are employed as pre-trained models and then fine-tuned on downstream tasks such as image classification, semantic segmentation, object detection, and action recognition etc. The performance of the transfer learning on these high-level vision tasks demonstrates the generalization ability of the learned features. If ConvNets of self-supervised learning can learn general features, then the pre-trained models can be used as a good starting point for other vision tasks that require capturing similar features from images or videos.

Image classification, semantic segmentation, and object detection usually are used as the tasks to evaluate the generalization ability of the learned image features by self-supervised learning methods, while human action recognition in videos is used to evaluate the quality of video features obtained from self-supervised learning methods. Below are brief introductions of the commonly used high-level tasks for visual feature evaluation.

\subsubsection{Semantic Segmentation}

Semantic segmentation, the task of assigning semantic labels to each pixel in images, is of great importance in many applications such as autonomous driving, human-machine interaction, and robotics. The community has recently made promising progress and various networks have been proposed such as Fully Convolutional Network (FCN) \cite{FCN}, DeepLab \cite{DeepLab}, PSPNet \cite{PSPNet} and datasets such as PASCAL VOC \cite{VOC}, CityScape \cite{Cityscape}, ADE20K \cite{ADE20K}. 

\begin{figure}[!ht]
\begin{center}
\includegraphics[width=0.5\textwidth]{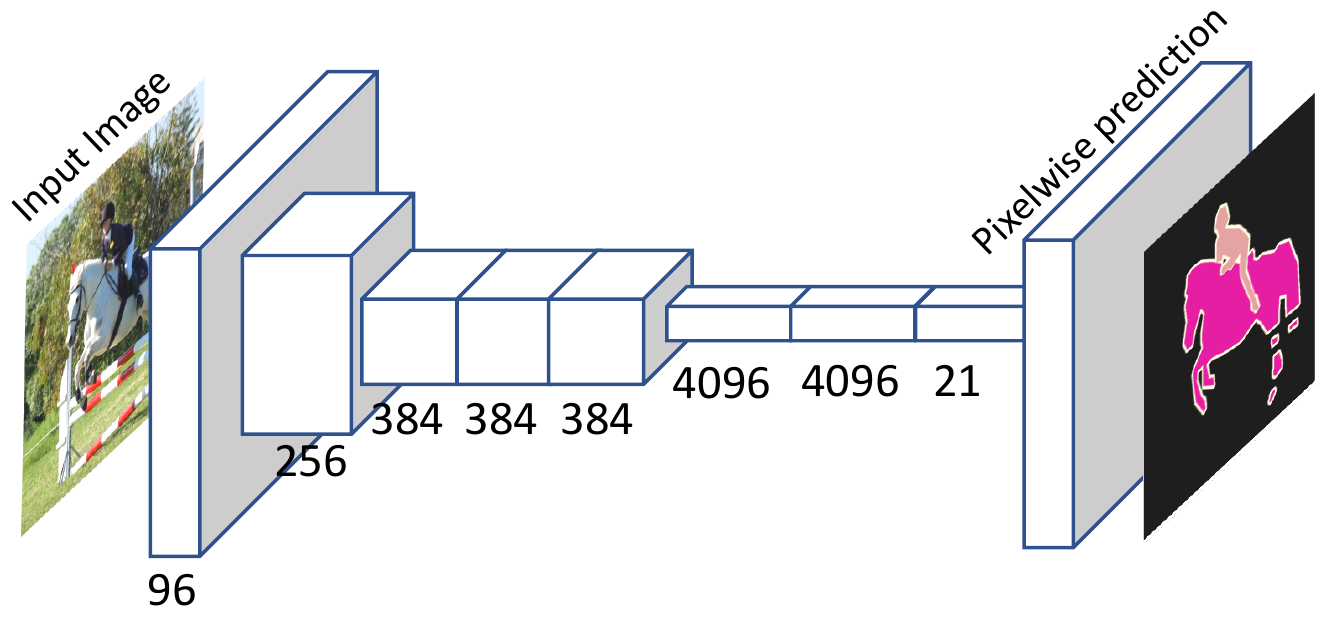}
\end{center}
\caption{The framework of the Fully Convolutional Neural Network proposed for semantic segmentation \cite{FCN}. Figure is reproduced based on \cite{FCN}.}
\label{fig:FCN}
\end{figure}

Among all these methods, FCN \cite{FCN} is a milestone work for semantic segmentation since it started the era of applying fully convolution network (FCN) to solve this task. The architecture of FCN is shown in Fig. \ref{fig:FCN}. 2DConvNet such as AlexNet, VGG, ResNet is used as the base network for feature extraction while the fully connected layer is replaced by transposed convolution layer to obtain the dense prediction. The network is trained end-to-end with pixel-wise annotations. 

When using semantic segmentation as downstream task to evaluate the quality of image features learned by self-supervised learning methods, the FCN is initialized with the parameters trained with the pretext task and fine-tuned on the semantic segmentation dataset, then the performance on the semantic segmentation task is evaluated and compared with that of other self-supervised methods.

\subsubsection{Object Detection}

Object Detection, a task of localizing the position of objects in images and recognizing the category of the objects, is also very import for many computer vision applications such as autonomous driving, robotics, scene text detection and so on. Recently, many datasets such as MSCOCO \cite{MSCOCO} and OpenImage \cite{OpenImage} have been proposed for object detection and many ConvNet-based models \cite{rcnn}, \cite{fastrcnn}, \cite{fasterrcnn}, \cite{YOLO}, \cite{YOLOv2}, \cite{SSD}, \cite{FPN}, \cite{RetinaNet} have been proposed and obtained great performance. 

\begin{figure}[!ht]
\begin{center}
\includegraphics[width=0.5\textwidth]{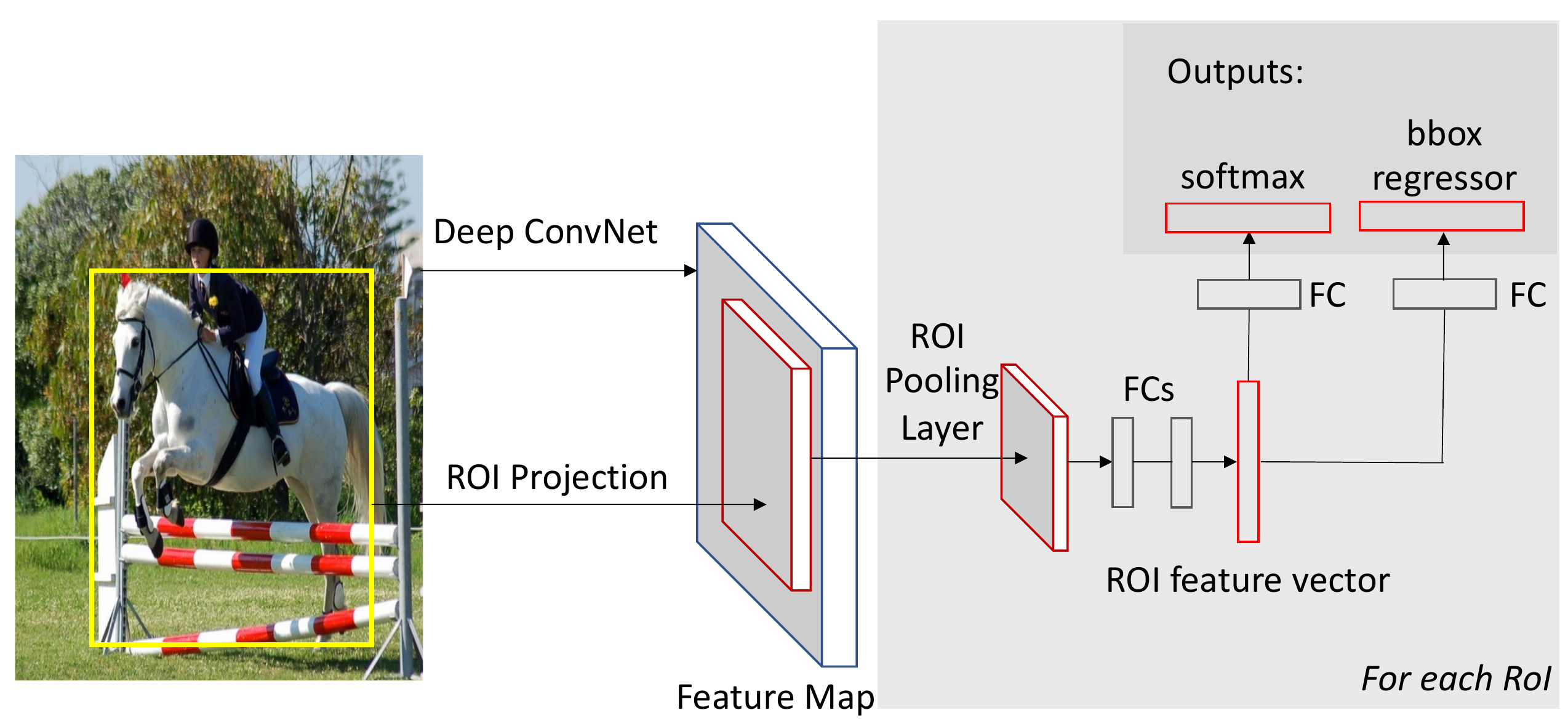}
\end{center}
\caption{The pipeline of the Fast-RCNN for object detection. Figure is reproduced based on \cite{fasterrcnn}.}
\label{fig:Faster-RCNN}
\end{figure}

Fast-RCNN \cite{fastrcnn} is a two-stage network for object detection. The framework of Fast-RCNN is shown in Fig.~\ref{fig:Faster-RCNN}. Object proposals are generated based on feature maps produced by a convolution neural network, then these proposals are fed to several fully connected layers to generate the bounding box of objects and the categories of these objects. 

When using object detection as downstream task to evaluate the quality of the self-supervised image features, networks that trained with the pretext task on unlabeled large data are served as the pre-trained model for the Fast-RCNN \cite{fastrcnn} and then fine-tuned on object detection datasets, then the performance on the object detection task is evaluated to demonstrate the generalization ability of self-supervised learned features.

\subsubsection{Image Classification}
Image Classification is a task of recognizing the category of objects in each image. Many networks have been designed for this task such as AlexNet \cite{AlexNet}, VGG \cite{VGG}, ResNet \cite{ResNet}, GoogLeNet \cite{GoogLeNet}, DenseNet \cite{DenseNet}, etc. Usually, only one class label is available for each image although the image may contains different classes of objects. 

When choosing image classification as a downstream task to evaluate the quality of image features learned from self-supervised learning methods, the self-supervised learned model is applied on each image to extract features which then are used to train a classifier such as Support Vector Machine (SVM) \cite{SVM}. The classification performance on testing data is compared with other self-supervised models to evaluate the quality of the learned features.

\subsubsection{Human Action Recognition}

Human action recognition is a task of identifying what people doing in videos for a list of pre-defined action classes. Generally, videos in human action recognition datasets contain only one action in each video \cite{UCF101, HMDB51, Kinetics}. Both the spatial and temporal features are needed to accomplish this task.

The action recognition task is often used to evaluate the quality of video features learned by self-supervised learning methods. The network is first trained on unlabeled video data with pretext tasks, then it is fine-tuned on action recognition datasets with human annotations to recognize the actions. The testing performance on action recognition task is compared with other self-supervised learning methods to evaluate the quality of the learned features.

\subsubsection{Qualitative Evaluation}

In addition to these quantitative evaluations of the learned features, there are also some qualitative visualization methods to evaluate the quality of self-supervised learning features. Three methods are often used for this purpose: kernel visualization, feature map visualization, and image retrieval visualization \cite{deepcluster, RotNet, 3DRotNet, contextprediction}.

\textbf{Kernel Visualization:} Qualitatively visualize the kernels of the first convolution layer learned with the pretext tasks and compare the kernels from supervised models. The similarity of the kernels learned by supervised and self-supervised models are compared to indicate the effectiveness of self-supervised methods \cite{deepcluster, 3DRotNet}.

\textbf{Feature Map Visualization:} Feature maps are visualized to show the attention of networks. Larger activation represents the neural network pays more attention to the corresponding region in the image. Feature maps are usually qualitatively visualized and compared with that of supervised models \cite{RotNet, 3DRotNet}.

\textbf{Nearest Neighbor Retrieval:} In general, images with similar appearance usually are closer in the feature space. The nearest neighbor method is used to find the top $K$ nearest neighbors from the feature space of the features learned by the self-supervised learned model \cite{shuffleandlearn, contextprediction, graphconstraint}.

\begin{table*}[!t]
    \centering
    \caption{Summary of commonly used image and video datasets. Note that image datasets can be used to learn image features, while video datasets can be used to learn both image and video features.}
    \begin{tabular}{|c|c|c|c|c|c|c|}
        \hline
         Dataset &Data Type &Size  &Synthetic & \# classes &Label\\
        \hline
        ImageNet       \cite{ImageNet}  &Image  &$1.3$ million images &\xmark &$1,000$ &Object category label\\
        Places   \cite{Places} &Image &$2.5$ million images &\xmark &$205$  &scene categories label\\
        Places365 \cite{Places365} &Image &$10$ million images &\xmark &$434$   &scene categories label\\
        SUNCG   \cite{SUNCG}       &Image &$150,000$ images  &\cmark &$84$    &depth, volumetric data\\
        MNIST \cite{MNIST}          &Image     &$70,000$ images      &\xmark & $10$    &Digit class label\\
        SVHN \cite{SVHN}             &Image     &$600,000$ Images      &\xmark & $10$    &Digit class label\\
        CIFAR10 \cite{CIFAR10}           &Image  &$60,000$ Images  &\xmark & $10$     &Object category label\\
        STL-10 \cite{STL-10}&Image  &$101,300$ Images  &\xmark & $10$     &Object category label\\
        PASCAL VOC \cite{VOC}   &Image   &$2,913$ images    &\xmark & $20$     &Category label, bounding box, segmentation mask\\
        \hline
        YFCC100M  \cite{YFCC100M} &Image/Video  &$100$ million media data &\xmark &---    &Hashtags\\
        SceneNet RGB-D \cite{SceneNetRGB-D}   &Video &$5$ million images      &\cmark     &$13$    &Depth, Instance Segmentation, Optical Flow\\
        Moment-in-Time  \cite{MITS}   &Video     &$1$ million 3-second videos   &\xmark & $339$    &Video category class\\
        Kinetics   \cite{Kinetics}        &Video     &$0.5$ million 10-second videos   &\xmark &$600$  &Human action class\\
        AudioSet \cite{AudioSet}      &Video     &$2$ million 10-second videos   &\xmark &$632$    &Audio event class \\
        KITTI \cite{KITTI} &Video  &$28$ videos  &\xmark    & ---  &Data captured by various sensors are available\\
        UCF101 \cite{UCF101}          &Video     &$10,031$ videos     &\xmark  & $101$  &Human action class\\
        HMDB51 \cite{HMDB51}            &Video     &$6,766$ videos      &\xmark    & $51$  &Human action class\\
        \hline
    \end{tabular}
    \label{tab:dataset}
\end{table*}

\section{Datasets}

This section summarizes the commonly used image and video datasets for training and evaluating of self-supervised visual feature learning methods. Self-supervised learning methods can be trained with images or videos by discarding human-annotated labels, therefore, any datasets that collected for supervised learning can be used for self-supervised visual feature learning without using human-annotated labels. The evaluation of the quality of learned features is normally conducted by fine-tuned on high-level vision tasks with relatively small datasets (normally with accurate labels) such as video action recognition, object detection, semantic segmentation, etc.  It is worth noting that networks use these synthetic datasets for visual feature learning are considered as self-supervised learning in this paper since labels of synthetic datasets are automatically generated by game engines and no human annotations are involved. Table \ref{tab:dataset} summarizes the commonly used image and video datasets.

\subsection{Image Datasets}

\begin{itemize}

  \item{\textbf{ImageNet:} The ImageNet dataset \cite{ImageNet} contains $1.3$ million images uniformly distributed into $1,000$ classes and is organized according to the WordNet hierarchy. Each image is assigned with only one class label. ImageNet is the most widely used dataset for self-supervised image feature learning.} 
 
  \item{\textbf{Places:} The Places dataset \cite{Places} is proposed for scene recognition and contains more than $2.5$ million images covering more than $205$ scene categories with more than $5,000$ images per category.}
  
  \item{\textbf{Places365:} The Places365 is the 2nd generation of the Places database which is built for high-level visual understanding tasks, such as scene context, object recognition, action and event prediction, and theory-of-mind inference \cite{Places365}. There are more than $10$ million images covering more than $400$ classes and $5,000$ to $30,000$ training images per class.}
 
  \item{\textbf{SUNCG:} The SUNCG dataset is a large synthetic 3D scene repository for indoor scenes which consists of over $45,000$ different scenes with manually created realistic room and furniture layouts \cite{SUNCG}. The synthetic depth, object level semantic labels, and volumetric ground truth are available.}

  \item{\textbf{MNIST: } The MNIST is a dataset of handwritten digits  consisting of $70,000$ images while $60,000$ images belong to training set and the rest $10,000$ images are for testing \cite{MNIST}. All digits have been size-normalized and centered in fixed-size images.} 
 
  \item{\textbf{SVHN: } SVHN is a dataset for recognizing digits and numbers in natural scene images which obtained from house numbers from Google Street View images \cite{SVHN}. The dataset consists of over $600,000$ images and all digits have been resized to a fixed resolution of $32 \times 32$ pixels.
  } 

  \item{\textbf{CIFAR10:} The CIFAR10 dataset is a collection of tiny images for image classification task \cite{CIFAR10}. It consists of $60,000$ images of size $32 \times 32$ that covers $10$ different classes. The $10$ classes include airplane, automobile, bird, cat, deer, dog, frog, horse, ship, and truck. The dataset is balanced and there are $6,000$ images of each class.}

  \item{\textbf{STL-10:} The STL-10 dataset is specifically designed for developing unsupervised feature learning \cite{STL-10}. It consists of $500$ labeled training images, $800$ testing images, and $100,000$ unlabeled images covering $10$ classes which include airplane, bird, car, cat, deer, dog, horse, monkey, ship, and truck.} 
 
  \item{\textbf{PASCAL Visual Object Classes (VOC):} The VOC 2,012 dataset \cite{VOC} contains $20$ object categories including vehicles, household, animals, and other: aeroplane, bicycle, boat, bus, car, motorbike, train, bottle, chair, dining table, potted plant, sofa, TV/monitor, bird, cat, cow, dog, horse, sheep, and person. Each image in this dataset has pixel-level segmentation annotations, bounding box annotations, and object class annotations. This dataset has been widely used as a benchmark for object detection, semantic segmentation, and classification tasks. The PASCAL VOC dataset is split into three subsets: $1,464$ images for training, $1,449$ images for validation and a private testing \cite{VOC}. All the self-supervised image representation learning methods are evaluated on this dataset with the three tasks.} 

%
\end{itemize}
 
\subsection{Video Datasets}

\begin{itemize}
  
  \item{\textbf{YFCC100M:} The Yahoo Flickr Creative Commons $100$ Million Dataset (YFCC100M) is a large public multimedia collection from Flickr, consisting of $100$ million media data, of which around $99.2$ million are images and $0.8$ million are videos \cite{YFCC100M}. The statistics on hashtags used in the YFCC100M dataset show that the data distribution is severely unbalanced \cite{YFCC100Munbalanced}.} 
  
  \item{\textbf{SceneNet RGB-D: } The SceneNet RGB-D dataset is a large indoor synthetic video dataset which consists of $5$ million rendered RGB-D images from over 15K trajectories in synthetic layouts with random but physically simulated object poses \cite{SceneNetRGB-D}. It provides pixel-level annotations for scene understanding problems such as semantic segmentation, instance segmentation, and object detection, and also for geometric computer vision problems such as optical flow, depth estimation, camera pose estimation, and 3D reconstruction \cite{SceneNetRGB-D}.}

  \item{\textbf{Moment in Time:} The Moment-in-Time dataset is a large balanced and diverse dataset for video understanding \cite{MITS}. The dataset consists of $1$ million video clips that cover $339$ classes, and each video lasts around $3$ seconds. The average number of video clips for each class is $1,757$ with a median of $2,775$. The video in this dataset contains videos that capturing visual and/or audible actions, produced by humans, animals, objects or nature \cite{MITS}.} 
 
  \item{\textbf{Kinetics:} The Kinetics dataset is a large-scale, high-quality dataset for human action recognition in videos \cite{Kinetics}. The dataset consists of around $500,000$ video clips covering $600$ human action classes with at least $600$ video clips for each action class. Each video clip lasts around 10 seconds and is labeled with a single action class.} 

  \item{\textbf{AudioSet:} The AudioSet consists of $2,084,320$ human-labeled 10-second sound clips drawn from YouTube videos covers ontology of $632$ audio event classes \cite{AudioSet}. The event classes cover a wide range of human and animal sounds, musical instruments and genres, and common everyday environmental sounds. This dataset is mainly used for the self-supervised learning from video and audio consistence \cite{AudioVisual}.} 

  \item{\textbf{KITTI:} The KITTI dataset is collected from driving a car around a city which equipped with various sensors including high-resolution RGB camera, gray-scale stereo camera, a 3D laser scanner, and high-precision GPS measurements and IMU accelerations from a combined GPS/IMU system \cite{KITTI}. Videos with various modalities captured by these sensors are available in this dataset.} 
  
  \item{\textbf{UCF101:} The UCF101 is a widely used video dataset for human action recognition \cite{UCF101}. The dataset consists of $13,370$ video clips with more than $27$ hours belonging to $101$ categories in this dataset. The videos in this dataset have a spatial resolution of $320\times240$ pixels and $25$ FPS frame rate. This dataset has been widely used for evaluating the performance of human action recognition. In the self-supervised sensorial, the self-supervised models are fine-tuned on the dataset and the accuracy of the action recognition are reported to evaluate the quality of the features.} 

  \item{\textbf{HMDB51:} Compared to other datasets, the HMDB51 dataset is a smaller video dataset for human action recognition. There are around $7,000$ video clips in this dataset belong to $51$ human action categories \cite{HMDB51}. The videos in HMDB51 dataset have $320\times240$ pixels spatial resolution and $30$ FPS frame rate. In the self-supervised sensorial, the self-supervised models are fine-tuned on the dataset to evaluate the quality of the learned video features.}

\end{itemize}

\section{Image Feature Learning}

In this section, three groups of self-supervised image feature learning methods are reviewed including generation-based methods, context-based methods, and free semantic label-based methods. A list of the image feature self-supervised learning methods can be found in Table \ref{tab:img-methods}. Since the cross modal-based methods mainly learn features from videos and most methods of this type can be used for both image and video feature learning, so cross modal-based methods are reviewed in the video feature learning section.

\begin{table*}[!t]
    \centering
    \caption{Summary of self-supervised image feature learning methods based on the category of pretext tasks. Multi-task means the method explicitly or implicitly uses multiple pretext tasks for image feature learning.}
    \begin{tabular}{|c|c|c|c|c|c|}
        \hline
         Method &Category &Code &Contribution\\
        \hline
        GAN \cite{GAN}       &Generation    &\cmark  &Forerunner of GAN\\
        DCGAN \cite{DCGAN}   &Generation    &\cmark  & Deep convolutional GAN for image generation\\
        WGAN  \cite{WGAN}    &Generation    &\cmark  &Proposed WGAN which makes the training of GAN more stable\\
        BiGAN  \cite{BiGAN}  &Generation    &\cmark  &Bidirectional GAN to project data into latent space\\
        SelfGAN  \cite{SelfGAN}  &Multiple            &\xmark  &Use rotation recognition and GAN for self-supervised learning\\
        ColorfulColorization \cite{colorfulcolorization}  &Generation   &\cmark  & Posing image colorization as a classification task \\
        Colorization \cite{colorproxy}   &Generation     
        &\cmark  &Using image colorization as the pretext task\\
        AutoColor \cite{AutoColor}  &Generation   &\cmark  &Training ConvNet to predict per-pixel color histograms \\        
        Split-Brain \cite{splitbrain} &Generation    &\cmark  &Using split-brain auto-encoder as the pretext task \\
        Context Encoder \cite{contextencoder} &Generation               &\cmark  &Employing ConvNet to solve image inpainting\\
        CompletNet \cite{CompletNet} &Generation     &\cmark  &Employing two discriminators to guarantee local and global consistent\\
        SRGAN \cite{SRGAN} &Generation    &\cmark  &Employing GAN for single image super-resolution\\
        SpotArtifacts \cite{SpotArtifacts} &Generation     &\cmark  &Learning by recognizing synthetic artifacts in images\\        
        \hline
        
        ImproveContext \cite{ImproveContext} &Context        &\xmark  & Techniques to improve context based self-supervised learning methods\\
        Context Prediction \cite{contextprediction}  &Context       &\cmark  &Learning by predicting the relative position of two patches from an image\\
        Jigsaw \cite{Jigsaw} &Context   &\cmark  &Image patch Jigsaw puzzle as the pretext task for self-supervised learning\\
        Damaged Jigsaw \cite{DamagedJigsaw}  &Multiple  &\xmark  &Learning by solving jigsaw puzzle, inpainting, and colorization together\\
        Arbitrary Jigsaw \cite{ArbitraryJigsaw} &Context  &\xmark  &Learning with jigsaw puzzles with arbitrary grid size and dimension \\
        DeepPermNet \cite{PermNet} &Context  &\cmark  &A new method to solve image patch jigsaw puzzle\\
        RotNet \cite{RotNet} &Context  &\cmark  & Learning by recognizing rotations of images\\
        Boosting \cite{boosting}    &Multiple     &\xmark  & Using clustering to boost the self-supervised learning methods\\
        JointCluster \cite{JointCluster}  &Context      &\cmark  & Jointly learning of deep representations and image clusters\\
        DeepCluster \cite{deepcluster}  &Context    &\cmark  & Using clustering as the pretext\\
        ClusterEmbegging \cite{ClusterEmbegging}  &Context    &\cmark  &Deep embedded clustering for self-supervised learning\\
        GraphConstraint \cite{graphconstraint}   &Context        &\cmark  &Learning with image pairs mined with Fisher Vector\\
        Ranking  \cite{wang2015unsupervised}  &Context   &\cmark  & Learning by ranking video frames with a triplet loss\\
        PredictNoise \cite{predictnoise} &Context    &\cmark  & Learning by mapping images to a uniform distribution over a manifold\\        
        MultiTask \cite{multitasklearning} &Multiple &\cmark  &  Using multiple pretext tasks for self-supervised feature learning\\   
        Learning2Count \cite{Learning2Count} &Context     &\cmark  & Learning by counting visual primitive\\ 
        \hline
        Watching Move \cite{watchingmove}   &Free Semantic Label        &\cmark  &Learning by grouping pixels of moving objects in videos\\        
        Edge Detection \cite{watchingmove}   &Free Semantic Label       &\cmark  &Learning by detecting edges\\
        Cross Domain \cite{watchingmove}   &Free Semantic Label         &\cmark  & Utilizing synthetic data and its labels rendered by game engines\\ 
        \hline
    \end{tabular}
    \label{tab:img-methods}
\end{table*}

\subsection{Generation-based Image Feature Learning}

Generation-based self-supervised methods for learning image features involve the process of generating images including image generation with GAN (to generate fake images), super-resolution (to generate high-resolution images), image inpainting (to predict missing image regions), and image colorization (to colorize gray-scale images into colorful images). For these tasks, pseudo training labels $P$ usually are the images themselves and no human-annotated labels are needed during training, therefore, these methods belong to self-supervised learning methods.

The pioneer work about the image generation-based methods is the Autoencoder \cite{Autoencoder} which learns to compress an image into a low-dimension vector which then is uncompressed into the image that closes to the original image with a bunch of layers. With an auto-encoder, networks can reduce the dimension of an image into a lower dimension vector that contains the main information of the original image. The current image generation-based methods follow a similar idea but with different pipelines to learn visual features through the process of image generation.

\subsubsection{Image Generation with GAN}

Generative Adversarial Network (GAN) is a type of deep generative model that was proposed by Goodfellow \textit{et al.} \cite{GAN}. A GAN model generally consists of two kinds of networks: a generator which is to generate images from latent vectors and a discriminator which is to distinguish whether the input image is generated by the generator. By playing the two-player game, the discriminator forces the generator to generate realistic images, while the generator forces the discriminator to improve its differentiation ability. During the training, the two networks are competing against with each other and make each other stronger.

\begin{figure}[!ht]
\begin{center}
\includegraphics[width=0.5\textwidth]{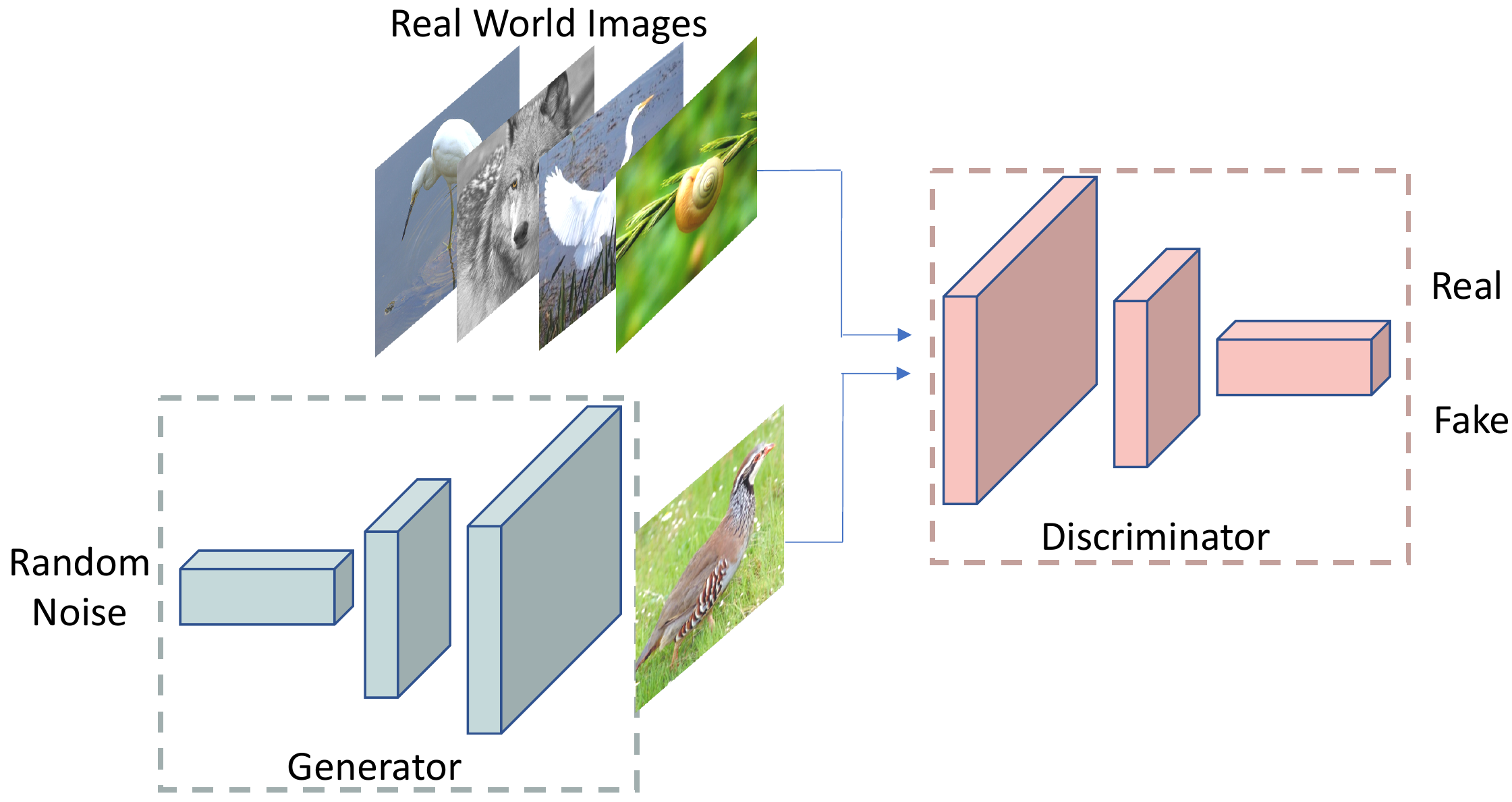}
\end{center}
\caption{The pipeline of Generative Adversarial Networks \cite{GAN}. By playing the two-player game, the discriminator forces the generator to generate realistic images, while the generator forces the discriminator to improve its differentiation ability.}
\label{fig:GAN}
\end{figure}

The common architecture for the image generation from a latent variable task is shown in Fig.~\ref{fig:GAN}. The generator is trained to map any latent vector sampled from latent space into an image, while the discriminator is forced to distinguish whether the image from the real data distribution or generated data distribution. Therefore, the discriminator is required to capture the semantic features from images to accomplish the task. The parameters of the discriminator can server as the pre-trained model for other computer vision tasks.

Mathematically, the generator $G$ is trained to learn a distribution $p_z$ of real word image data to generate realist data that undistinguished from the real data, while the discriminator $D$ is trained to distinguish the distribution of the real data $p_{data}$ and of the data distribution $p_z$ generated by the generator $G$. The min-max game between the generator $G$ and the discriminator $D$ is formulated as:

\begin{equation}
\min_{G}\max_{D} \mathbb{E}_{x \sim p_{data}(x)[log D(x)]} + \mathbb{E}_{z\sim p_{z}(z)[log(1-D(G(z)))]},
\label{eq:gan}
\end{equation}where $x$ is the real data, $G(z)$ is the generated data.

The discriminator $D$ is trained to maximize the probability for the real data $x$ (that is, $\mathbb{E}_{x \sim p_{data}(x)[log D(x)]}$) and minimize the probability for the generated data $G(z)$ (that is, $\mathbb{E}_{x \sim p_{data}(x)[log D(x)]}$). The generator is trained to generate data that close to real data $x$, so as the output of the discriminator is maximized $\mathbb{E}_{x \sim p_{data}(x)[log D(G(z))]}$. 

Most of the methods for image generation from random variables do not need any human-annotated labels. However, the main purpose of this type of task is to generate realistic images instead of obtaining better performance on downstream applications. Generally, the inception scores of the generated images are used to evaluate the quality of the generated images \cite{ImprovedGAN, FID}. And only a few methods evaluated the quality of the feature learned by the discriminator on the high-level tasks and compared with others \cite{SelfGAN, BiGAN, DCGAN}.

The adversarial training can help the network to capture the real distribution of the real data and generate realists data, and it has been widely used in computer vision tasks such as image generation \cite{BigGAN, StyleGAN}, video generation \cite{VideoGAN},\cite{MocoGAN}, super-resolution \cite{SRGAN}, image translation \cite{pix2pix}, and image inpainting \cite{CompletNet, contextencoder}. When there is no human-annotated label involves, the method falls into the self-supervised learning.

\subsubsection{Image Generation with Inpainting}


\begin{figure}[!ht]
\begin{center}
\includegraphics[width=0.45\textwidth]{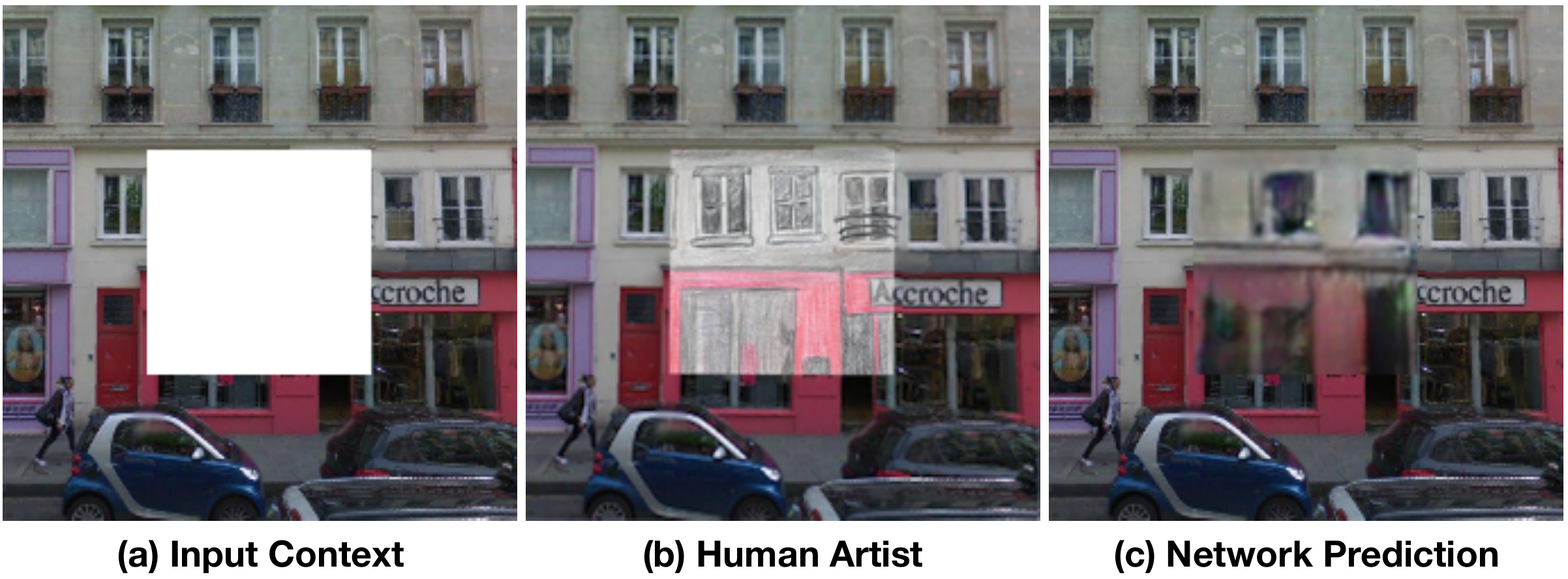}
\end{center}
\caption{Qualitative illustration of image inpainting task. Given an image with a missing region (a), a human artist has no trouble inpainting it (b). Automatic inpainting using context encoder proposed in \cite{contextencoder} trained with L2 reconstruction loss and adversarial loss is shown in (c). Figure is reproduced based on \cite{contextencoder}.}
\label{fig:Inpainting}
\end{figure}

Image inpainting is a task of predicting arbitrary missing regions based on the rest of an image. A qualitative illustration of the image inpainting task is shown in Fig.~\ref{fig:Inpainting}. The Fig.~\ref{fig:Inpainting}(a) is an image with a missing region, while the Fig~\ref{fig:Inpainting}(c) is the prediction of networks. To correctly predict missing regions, networks are required to learn the common knowledge including the color and structure of the common objects. Only by knowing this knowledge, networks are able to infer missing regions based on the rest part of the image.

By analogy with auto-encoders, Pathak \textit{et al.} made the first step to train a ConvNet to generate the contents of an arbitrary image region based on the rest of the image \cite{contextencoder}. Their contributions are in two folds: using a ConvNet to tackle image inpainting problem, and using the adversarial loss to help the network generate a realistic hypothesis. Most of the recent methods follow a similar pipeline \cite{CompletNet}. Usually, there are two kinds of networks: a generator network is to generate the missing region with the pixel-wise reconstruction loss and a discriminator network is to distinguish whether the input image is real with an adversarial loss. With the adversarial loss, the network is able to generate sharper and realistic hypothesis for the missing image region. Both the two kinds of networks are able to learn the semantic features from images and can be transferred to other computer vision tasks. However, only Pathak \textit{et al.} \cite{contextencoder} studied the performance of transfer learning for the learned parameters of the generator from the image inpainting task. 

The generator network which is a fully convolutional network has two parts: encoder and decoder. The input of the encoder is the image that needs to be inpainted and the context encoder learns the semantic feature of the image. The context decoder is to predict the missing region based on this feature. The generator is required to understand the content of the image in order to generate a plausible hypothesis. The discriminator is trained to distinguish whether the input image is the output of the generator. To accomplish the image inpainting task, both networks are required to learn semantic features of images.

\subsubsection{Image Generation with Super Resolution}

Image super-resolution (SR) is a task of enhancing the resolution of images. With the help of fully convolutional networks, finer and realistic high-resolution images can be generated from low-resolution images. SRGAN is a generative adversarial network for single image super-resolution proposed by Ledig et al. \cite{SRGAN}. The insight of this approach is to take advantage of the perceptual loss which consists of an adversarial loss and a content loss. With the perceptron loss, the SRGAN is able to recover photo-realistic textures from heavily downsampled images and show significant gains in perceptual quality.

There are two networks: one is generator which is to enhance the resolution of the input low-resolution image and the other is the discriminator which is to distinguish whether the input image is the output of the generator. The loss function for the generator is the pixel-wise L2 loss plus the content loss which is the similarity of the feature of the predicted high-resolution image and the high-resolution original image, while the loss for the discriminator is the binary classification loss. Compared to the network that only minimizing the Mean Squared Error (MSE) which generally leads to high peak signal-to-noise ratios but lacking high-frequency details, the SRGAN is able to recover the fine details of the high-resolution image since the adversarial loss pushes the output to the natural image manifold by the discriminator network. 

The networks for image super-resolution task are able to learn the semantic features of images. Similar to other GANs, the parameters of the discriminator network can be transferred to other downstream tasks. However, no one tested the performance of the transferred learning on other tasks yet. The quality of the enhanced image is mainly compared to evaluate the performance of the network.

\subsubsection{Image Generation with Colorization}

\begin{figure}[!ht]
\begin{center}
\includegraphics[width=0.5\textwidth]{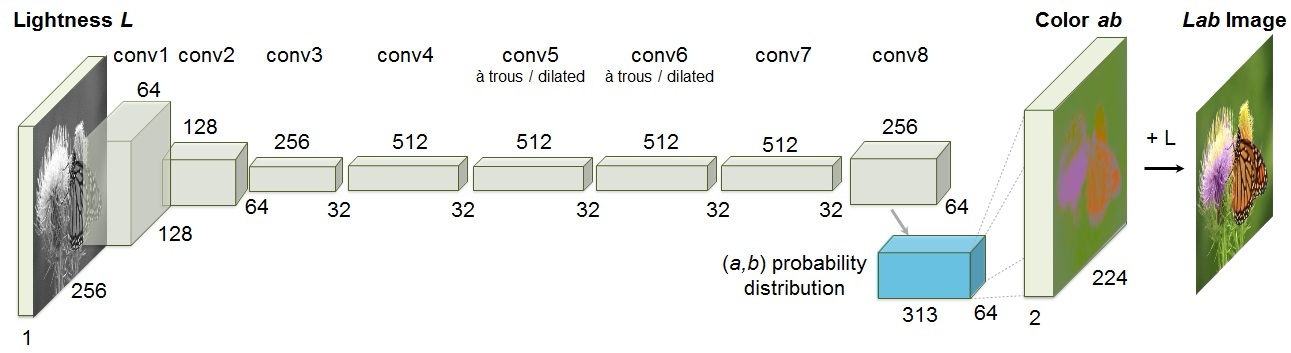}
\end{center}
\caption{The architecture of image colorization proposed in \cite{colorfulcolorization}. The figure is from \cite{colorfulcolorization} with author's permission.}
\label{fig:colorization}
\end{figure}

Image colorization is a task of predicting a plausible color version of the photograph given a gray-scale photograph as input. A qualitative illustration of the image colorization task is shown in Fig.~\ref{fig:colorization}. To correctly colorize each pixel, networks need to recognize objects and to group pixels of the same part together. Therefore, visual features can be learned in the process of accomplishing this task.

Many deep learning-based colorization methods have been proposed in recent years \cite{colorfulcolorization, InteractiveColorization, Letherebecolor}. A straightforward idea would be to employ a fully convolution neural network which consists of an encoder for feature extraction and a decoder for the color hallucination to colorization. The network can be optimized with L2 loss between the predicted color and its original color. Zhang \textit{et al.} proposed to handle the uncertainty by posting the task as a classification task and used class-rebalancing to increase the diversity of predicted colors \cite{colorfulcolorization}. The framework for image colorization proposed by Zhang \textit{et al.} is shown in Fig.~\ref{fig:colorization}. Trained in large-scale image collections, the method shows great results and fools human on $32$\% of the trials during the colorization test.

Some work specifically employs the image colorization task as the pretext for self-supervised image representation learning \cite{colorproxy, colorfulcolorization, AutoColor, splitbrain}. After the image colorization training is finished, the features learned through the colorization process are specifically evaluated on other downstream high-level tasks with transfer learning.

\subsection{Context-Based Image Feature Learning}

The context-based pretext tasks mainly employ the context features of images including context similarity, spatial structure, and temporal structure as the supervision signal. Features are learned by ConvNet through the process of solving the pretext tasks designed based on attributes of the context of images.

\subsubsection{Learning with Context Similarity}

\begin{figure}[!ht]
\begin{center}
\includegraphics[width=0.5\textwidth]{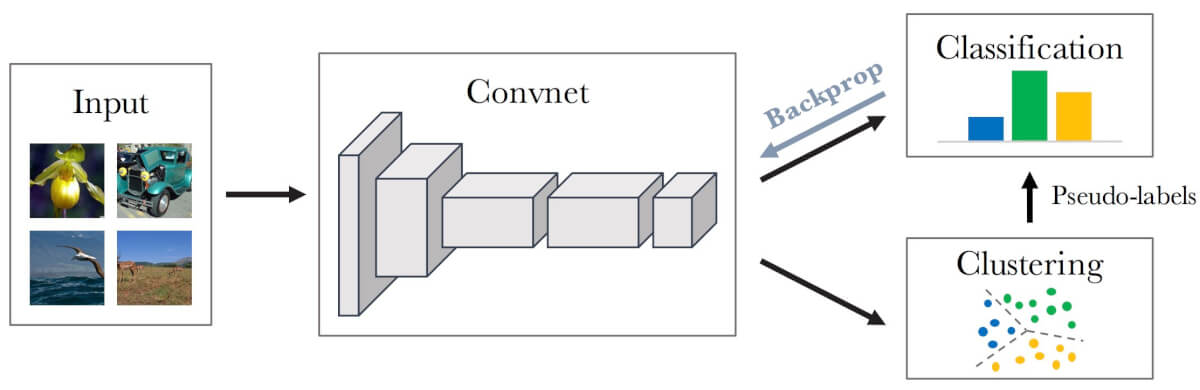}
\end{center}
\caption{The architecture of DeepClustering \cite{deepcluster}. The features of images are iteratively clustered and the cluster assignments are used as pseudo-labels to learn the parameters of the ConvNet. The figure is from \cite{deepcluster} with author's permission.}
\label{fig:clustering}
\end{figure}

Clustering is a method of grouping sets of similar data in the same clusters. Due to its powerful ability of grouping data by using the attributes of the data, it is widely used in many fields such as machine learning, image processing, computer graphics, etc. Many classical clustering algorithms have been proposed for various applications \cite{ClusteringReview}.

In the self-supervised scenario, the clustering methods mainly employed as a tool to cluster image data. A naive method would be to cluster the image data based on the hand-designed feature such as HOG \cite{HOG}, SIFT \cite{SIFT}, or Fisher Vector \cite{FisherVector}. After the clustering, several clusters are obtained while the image within one cluster has a smaller distance in feature space and images from different clusters have a larger distance in feature space. The smaller the distance in feature space, the more similar the image in the appearance in the RGB space. Then a ConvNet can be trained to classify the data by using the cluster assignment as the pseudo class label. To accomplish this task, the ConvNet needs to learn the invariance within one class and the variance among different classes. Therefore, the ConvNet is able to learn semantic meaning of images.

The existing methods about using the clustering variants as the pretext task follow these principals \cite{deepcluster, graphconstraint, boosting, JointCluster, ClusterEmbegging}. Firstly, the image is clustered into different clusters which the images from the same cluster have smaller distance and images from different clusters have larger distance. Then a ConvNet is trained to recognize the cluster assignment \cite{deepcluster, boosting} or to recognize whether two imaged are from same cluster \cite{graphconstraint}. The pipeline of DeepCluster, a clustering based methods, is shown in Fig.~\ref{fig:clustering}. DeepCluster iteratively clusters images with Kmeans and use the subsequent assignments as supervision to update the weights of the network. And it is the current state-of-the-art for the self-supervised image representation learning.

\subsubsection{Learning with Spatial Context Structure}

Images contain rich spatial context information such as the relative positions among different patches from an image which can be used to design the pretext task for self-supervised learning. The pretext task can be to predict the relative positions of two patches from same image \cite{contextprediction}, or to recognize the order of the shuffled a sequence of patches from same image \cite{Jigsaw, DamagedJigsaw, ArbitraryJigsaw}. The context of full images can also be used as a supervision signal to design pretext tasks such as to recognize the rotating angles of the whole images \cite{RotNet}. To accomplish these pretext tasks, ConvNets need to learn spatial context information such as the shape of the objects and the relative positions of different parts of an object.

\textbf{\begin{figure}[!ht]
\begin{center}
\includegraphics[width=0.5\textwidth]{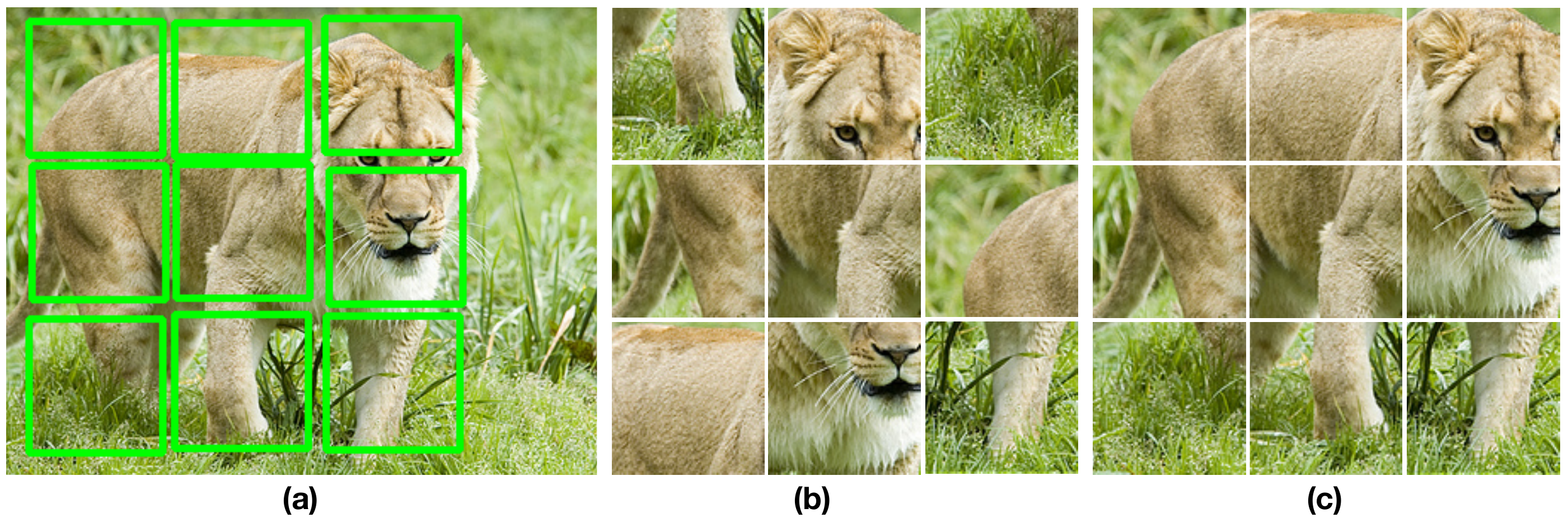}
\end{center}
\caption{The visualization of the Jigsaw Image Puzzle \cite{Jigsaw}. (a) is an image with $9$ sampled image patches, (b) is an example of shuffled image patches, and (c) shows the correct order of the sampled $9$ patches. Figure is reproduced based on \cite{Jigsaw}.}
\label{fig:Jigsaw}
\end{figure}}

The method proposed by Doersch \textit{et al.} is one of the pioneer work of using spatial context cues for self-supervised visual feature learning \cite{contextprediction}. Random pairs of image patches are extracted from each image, then a ConvNet is trained to recognize the relative positions of the two image patches. To solve this puzzle, ConvNets need to recognize objects in images and learn the relationships among different parts of objects. To avoid the network learns trivial solutions such as simply using edges in patches to accomplish the task, heavy data augmentation is applied during the training phase.  

Following this idea, more methods are proposed to learn image features by solving more difficult spatial puzzles \cite{Jigsaw, DamagedJigsaw, ArbitraryJigsaw, VideoJigsaw, CubicPuzzles}. As illustrated in Fig.~\ref{fig:Jigsaw}, one typical work proposed by Noroozi \textit{et al.} attempted to solve an image Jigsaw puzzle with ConvNet \cite{Jigsaw}. Fig.~\ref{fig:Jigsaw}(a) is an image with $9$ sampled image patches, Fig~\ref{fig:Jigsaw}(b) is an example of shuffled image patches, and Fig~\ref{fig:Jigsaw}(c) shows the correct order of the sampled $9$ patches. The shuffled image patches are fed to the network which trained to recognize the correct spatial locations of the input patches by learning spatial context structures of images such as object color, structure, and high-level semantic information. 

Given $9$ image patches from an image, there are $362,880$ $(9!)$  possible permutations and a network is very unlikely to recognize all of them because of the ambiguity of the task. To limit the number of permutations, usually, hamming distance is employed to choose only a subset of permutations among all the permutations that with relative large hamming distance. Only the selected permutations are used to train ConvNet to recognize the permutation of shuffled image patches \cite{ArbitraryJigsaw, DamagedJigsaw, RL, Jigsaw}.

The main principle of designing puzzle tasks is to find a suitable task which is not too difficult and not too easy for a network to solve. If it is too difficult, the network may not converge due to the ambiguity of the task or can easily learn trivial solutions if it is too easy. Therefore, a reduction in the search space is usually employed to reduce the difficulty of the task. 

\subsection{Free Semantic Label-based Image Feature Learning}

The free semantic label refers to labels with semantic meanings that obtained without involving any human annotations. Generally, the free semantic labels such as segmentation masks, depth images, optic flows, and surface normal images can be rendered by game engine or generated by hard-code methods. Since these semantic labels are automatically generated, the methods using the synthetic datasets or using them in conjunction with a large unlabeled image or video datasets are considered as self-supervised learning methods.

\subsubsection{Learning with Labels Generated by Game Engines}

\begin{figure}[!ht]
\begin{center}
\includegraphics[width=0.5\textwidth]{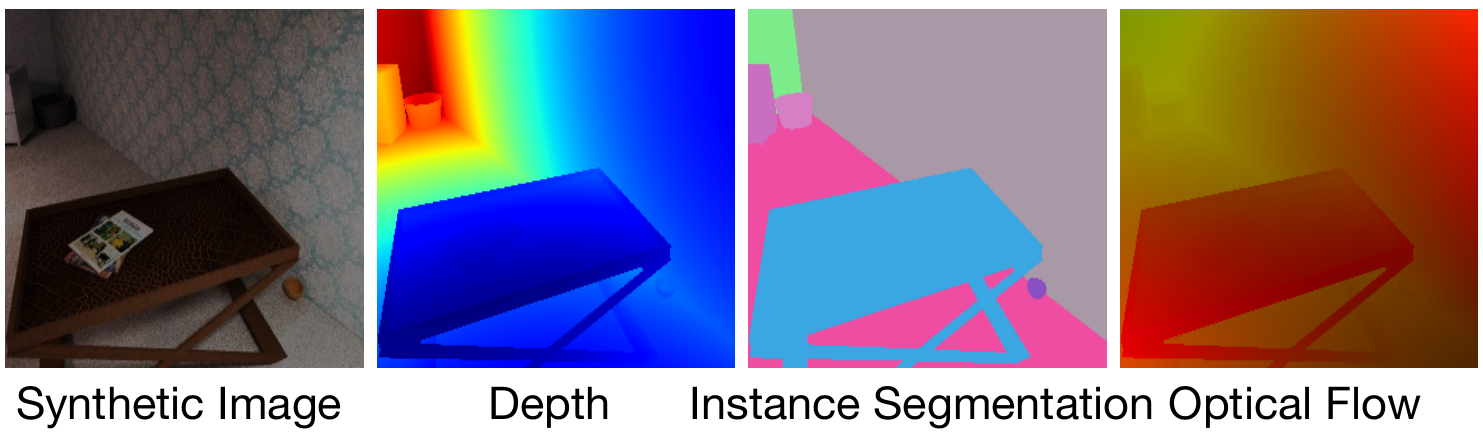}
\end{center}
\caption{An example of an indoor scene generated by a game engine \cite{SceneNetRGB-D}. For each synthetic image, the corresponding depth, instance segmentation, and optical flow can be automatically generated by the engine.}
\label{fig:synthetic}
\end{figure}

Given models of various objects and layouts of environments, game engines are able to render realistic images and provide accurate pixel-level labels. Since game engines can generate large-scale datasets with negligible cost, various game engines such as Airsim \cite{Airsim} and Carla \cite{CARLA} have been used to generate large-scale synthetic datasets with high-level semantic labels including depth, contours, surface normal, segmentation mask, and optical flow for training deep networks. An example of an RGB image with its generated accurate labels is shown in Fig.~\ref{fig:synthetic}. 

Game engines can generate realistic images with accurate pixel-level labels with very low cost. However, due to the domain gap between synthetic and real-world images, the ConvNet purely trained on synthetic images cannot be directly applied to real-world images. To utilize synthetic datasets for self-supervised feature learning, the domain gap needs to be explicitly bridged. In this way, the ConvNet trained with the semantic labels of the synthetic dataset can be effectively applied to real-world images.

\begin{figure}[!ht]
\begin{center}
\includegraphics[width=0.5\textwidth]{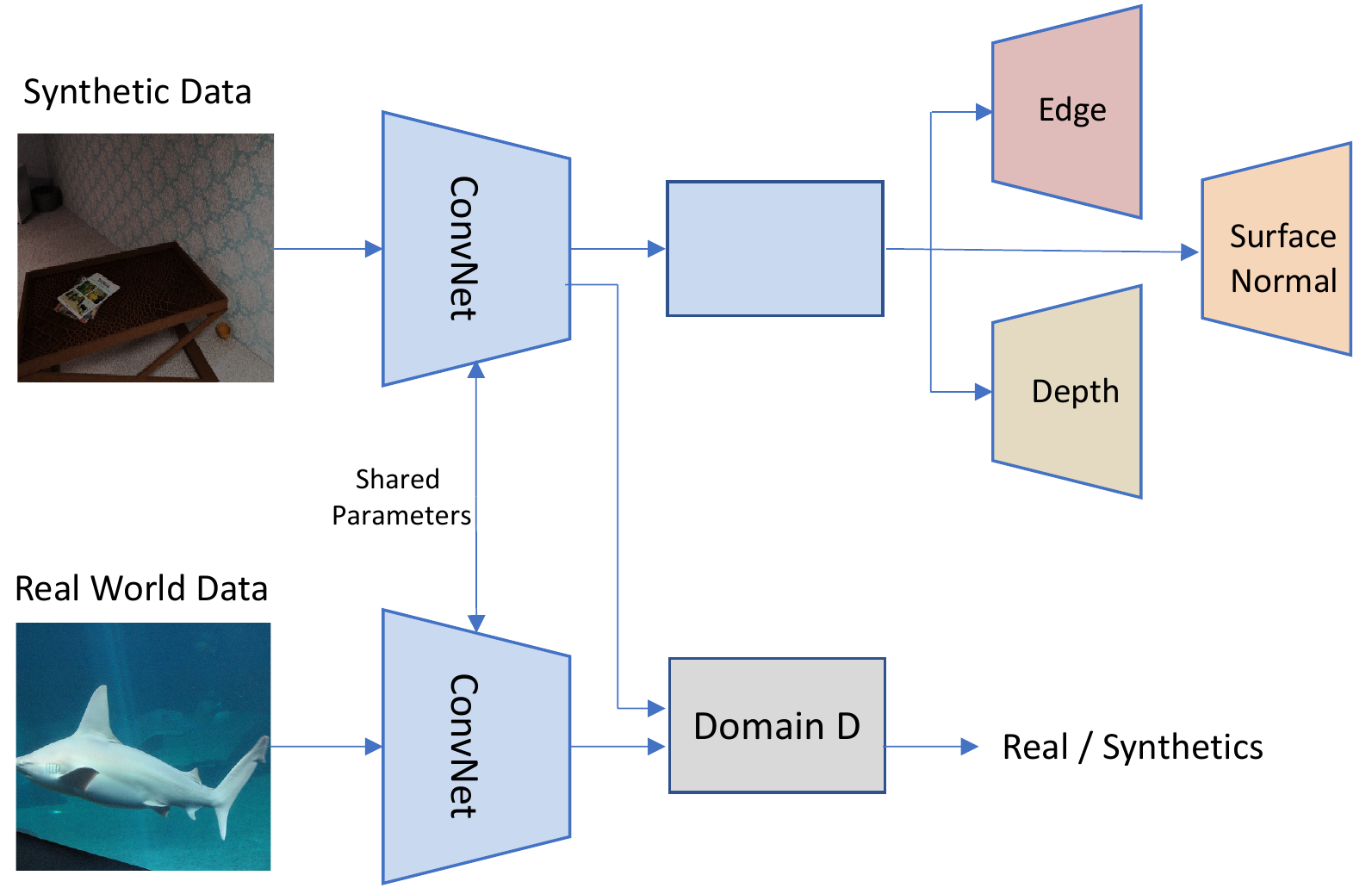}
\end{center}
\caption{The architecture for utilizing synthetic and real-world images for self-supervised feature learning \cite{crossdomain}. Figure is reproduced based on \cite{crossdomain}.}
\label{fig:crossdomain}
\end{figure}

To overcome the problem, Ren and Lee proposed an unsupervised feature space domain adaptation method based on adversarial learning \cite{crossdomain}. As shown in Fig.~\ref{fig:crossdomain}, the network predicts surface normal, depth, and instance contour for the synthetic images and a discriminator network $D$ is employed to minimize the difference of feature space domains between real-world and synthetic data. Helped with adversarial training and accurate semantic labels of synthetic images, the network is able to capture visual features for real-world images.

Compared to other pretext tasks in which the pretext tasks implicitly force ConvNets to learn semantic features, this type of methods are trained with accurate semantic labels which explicitly force ConvNets to learn features that highly related to the objects in images.

\subsubsection{Learning with Labels Generated by Hard-code programs}

Applying hard-code programs is another way to automatically generate semantic labels such as salience, foreground masks, contours, depth for images and videos. With these methods, very large-scale datasets with generated semantic labels can be used for self-supervised feature learning. This type of methods generally has two steps: (1) label generation by employing hard-code programs on images or videos to obtain labels, (2) train ConvNets with the generated labels.

Various hard-code programs have been applied to generate labels for self-supervised learning methods include methods for foreground object segmentation \cite{watchingmove}, edge detection \cite{unsupervisededges}, and relative depth prediction \cite{relativedepth}. Pathak \textit{et al.} proposed to learn features by training a ConvNet to segment foreground objects in each frame of a video while the label is the mask of moving objects in videos \cite{watchingmove}. Li \textit{et al.} proposed to learn features by training a ConvNet for edge prediction while labels are motion edges obtained from flow fields from videos \cite{unsupervisededges}. Jing \textit{et al.} proposed to learn features by training a ConvNet to predict relative scene depths while the labels are generated from optical flow \cite{relativedepth}. 

No matter what kind of labels used to train ConvNets, the general idea of this type of methods is to distill knowledge from hard-code detector. The hard-code detector can be edge detector, salience detector, relative detector, etc. As long as no human-annotations are involved through the design of detectors, then the detectors can be used to generate labels for self-supervised training.

Compared to other self-supervised learning methods, the supervision signal in these pretext tasks is semantic labels which can directly drive the ConvNet to learn semantic features. However, one drawback is that the semantic labels generated by hard-code detector usually are very noisy which need to specifically cope with.

\section{Video Feature Learning}


This section reviews the self-supervised methods for learning video features, as listed in Table \ref{tab:vid-methods}, they can be categorized into four classes:  generation-based methods, context-based methods, free semantic label-based methods, and cross modal-based methods.

Since video features can be obtained by various kinds of networks including 2DConvNet, 3DConvNet, and LSTM combined with 2DConvNet or 3DConvNet. When 2DConvNet is employed for video self-supervised feature learning, then the 2DConvNet is able to extract both image and video features after the self-supervised pretext task training finished.

\begin{table*}[!t]
    \centering
    \caption{Summary of self-supervised video feature learning methods based on the category of pretext tasks.}
    \begin{tabular}{|c|c|c|c|c|c|}
        \hline
         Mehtod &SubCategory &Code &Contribution\\
        \hline
        VideoGAN \cite{VideoGAN}   &Generation                  &\cmark  & Forerunner of video generation with GAN\\
        MocoGAN \cite{MocoGAN}   &Generation             &\cmark  & Decomposing motion and content for video generation with GAN\\
        TemporalGAN \cite{TemporalGAN}   &Generation          &\cmark  & Decomposing temporal and image generator for video generation\\
        Video Colorization \cite{videocolorize}  &Generation     &\cmark  &  Employing video colorization as the pretext task\\
        Un-LSTM  \cite{Self-LSTM}     &Generation             &\cmark  &  Forerunner of video prediction with LSTM\\
        ConvLSTM \cite{ConvLSTM}     &Generation    &\cmark  & Employing Convolutional LSTM for video prediction\\
        MCNet \cite{McNet}    &Generation    &\cmark  & Disentangling motion and content for video prediction\\
        LSTMDynamics \cite{LSTMDynamics}   &Generation    &\xmark  &  Learning by predicting long-term temporal dynamic in videos\\
        \hline
        Video Jigsaw \cite{VideoJigsaw}    &Context  
        &\xmark  & Learning by jointly reasoning about spatial and temporal context\\
        Transitive \cite{transitive}  &Context    &\xmark  & Learning inter and intra instance variations with a Triplet loss\\
        3DRotNet \cite{3DRotNet} &Context        &\xmark  &  Learning by recognizing rotations of video clips\\
        CubicPuzzles \cite{CubicPuzzles}  &Context  &\xmark  &   Learning by solving video cubic puzzles\\
        ShuffleLearn \cite{shuffleandlearn}    &Context             &\cmark & Employing temporal order verification as the pretext task\\
        LSTMPermute \cite{LSTMPermute}    &Context    &\cmark  &  Learning by temporal order verification with LSTM\\
        OPN \cite{sortsequence}    &Context        &\cmark  &   Using frame sequence order recognition as the pretext task\\
        O3N  \cite{O3N}  &Context  &\xmark  &  Learning by identifying odd video sequences\\        
        ArrowTime \cite{arrowoftime}  &Context      &\cmark  &  Learning by recognizing the arrow of time in videos\\
        TemporalCoherence \cite{TemporalCoherence}    &Context    &\xmark  &  Learning with the temporal coherence of features of frame sequence\\
    
        \hline
        FlowNet \cite{FlowNet}      &Cross Modal            &\cmark  & Forerunner of optical flow estimation with ConvNet\\
        FlowNet2 \cite{FlowNet2}      &Cross Modal            &\cmark  & Better architecture and better performance on optical flow estimation\\
        UnFlow \cite{UnFlow}      &Cross Modal      &\cmark  &  An unsupervised loss for optical flow estimation\\
    
        CrossPixel \cite{crosspixel} &Cross Modal        &\xmark  &  Learning by predicting motion from a single image as the pretext task\\
        CrossModel \cite{crossmodel} &Cross Modal        &\xmark  &  Optical flow and RGB correspondence verification as pretext task\\
        AVTS \cite{AVTS}    &Cross Modal        &\xmark  & Visual and Audio correspondence verification as pretext task\\
        AudioVisual \cite{AudioVisual}    &Cross Modal      &\cmark  &  Jointly modeling visual and audio as fused multisensory representation\\
        LookListenLearn \cite{looklistenlearn}  &Cross Modal         &\cmark  &  Forerunner of Audio-Visual Correspondence for self-supervised learning \\
        AmbientSound \cite{AmbientSound}  &Cross Modal  &\xmark  & Predicting a statistical summary of the sound from a video frame\\  
        EgoMotion \cite{egomotion}  &Cross Modal    &\cmark  
        & Learning by predicting camera motion and the scene structure from videos\\  
        LearnByMove \cite{learn2seebymove}  &Cross Modal     &\cmark  &  Learning by predicting the camera transformation from a pairs of images\\  
        TiedEgoMotion \cite{TiedEgoMotion}  &Cross Modal     &\xmark  &  Learning from ego-motor signals and video sequence\\ 
        GoNet \cite{GeoNet}  &Cross Modal     &\cmark  &  Jointly learning monocular depth, optical flow and ego-motion estimation from videos\\  
        DepthFlow \cite{DepthFlow}  &Cross Modal     &\cmark  &Depth and optical flow learning using cross-task consistency from videos \\ 
        VisualOdometry \cite{VisualOdometry}  &Cross Modal     &\cmark  & An unsupervised paradigm for deep visual odometry learning\\
        ActivesStereoNet \cite{ActivesStereoNet}  &Cross Modal  &\cmark  & End-to-end self-supervised learning of depth from active stereo systems\\
        \hline
    \end{tabular}
    \label{tab:vid-methods}
\end{table*}

\subsection{Generation-based Video Feature Learning }

Learning from video generation refers to the methods that visual features are learned through the process of video generation while without using any human-annotated labels. This type of methods includes video generation with GAN \cite{VideoGAN}, video colorization \cite{videocolorize} and video prediction \cite{Self-LSTM}. For these pretext tasks, the pseudo training label $P$ usually is the video itself and no human-annotated labels are needed during training, therefore, these methods belong to self-supervised learning.

\subsubsection{Learning from Video Generation}

\begin{figure}[!ht]
\begin{center}
\includegraphics[width=0.5\textwidth]{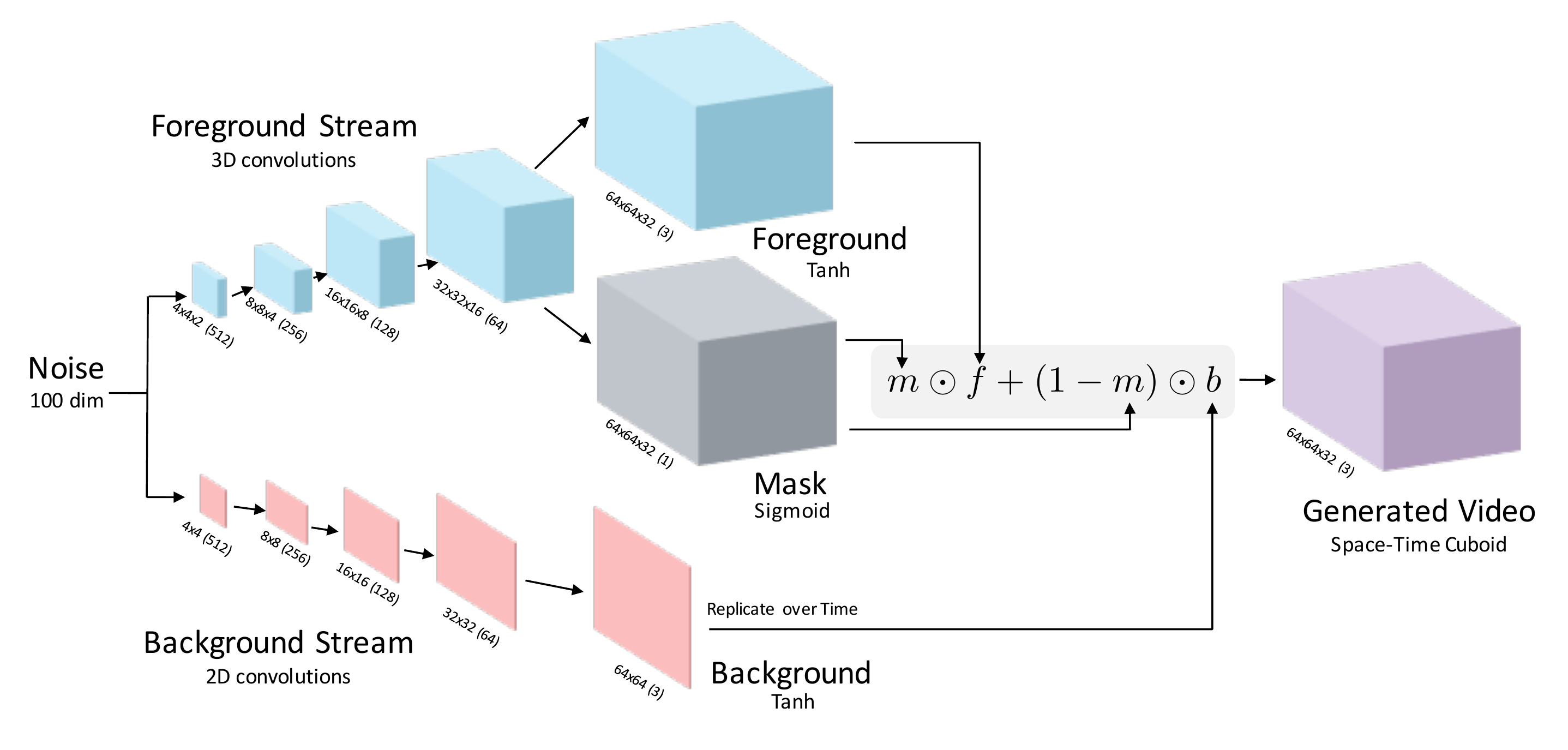}
\end{center}
\caption{The architecture of the generator in VideoGan for video generation with GAN proposed in \cite{VideoGAN}. The figure is from \cite{VideoGAN} with author's permission.}
\label{fig:VideoGAN}
\end{figure}

After GAN-based methods obtained breakthrough results in image generation, researchers employed GAN to generate videos \cite{MocoGAN, VideoGAN, TemporalGAN}. One pioneer work of video generation with GAN is VideoGAN \cite{VideoGAN}, and the architecture of the generator network is shown in Fig.~\ref{fig:VideoGAN}. To model the motion of objects in videos, a two-stream network is proposed for video generation while one stream is to model the static regions in in videos as background and another stream is to model moving object in videos as foreground \cite{VideoGAN}. Videos are generated by the combination of the foreground and background streams. The underline assumption is that each random variable in the latent space represents one video clip. This method is able to generate videos with dynamic contents. However, Tulyakov \textit{et al.} argues that this assumption increases difficulties of the generation, instead, they proposed MocoGAN to use the combination of two subspace to represent a video by disentangling the context and motions in videos \cite{MocoGAN}. One space is context space which each variable from this space represents one identity, and another space is motion space while the trajectory in this space represents the motion of the identity. With the two sub-spaces, the network is able to generate videos with higher inception score.

The generator learns to map latent vectors from latent space into videos, while discriminator learns to distinguish the real world videos with generated videos. Therefore, the discriminator needs to capture the semantic features from videos to accomplish this task. When no human-annotated labels are used in these frameworks, they belong to the self-supervised learning methods. After the video generation training on large-scale unlabeled dataset finished, the parameters of discriminator can be transferred to other downstream tasks \cite{VideoGAN}.

\subsubsection{Learning from Video Colorization}

Temporal coherence in videos refers to that consecutive frames within a short time have similar coherent appearance. The coherence of color can be used to design pretext tasks for self-supervised learning. One way to utilize color coherence is to use video colorization as a pretext task for self-supervised video feature learning. 

Video colorization is a task to colorize gray-scale frames into colorful frames. Vondrick \textit{et al.} proposed to constrain colorization models to solve video colorization by learning to copy colors from a reference frame \cite{videocolorize}. Given the reference RGB frame and a gray-scale image, the network needs to learn the internal connection between the reference RGB frame and gray-scale image to colorize it.

Another perspective is to tackle video colorization by employing a fully convolution neural network. Tran \textit{et al.} proposed an U-shape convolution neural network for video colorization \cite{voxel2voxel}. The network is an encoder-decoder based 3DConvNet. The input of the network is a clip of grayscale video clip, while the output if a colorful video clip. The encoder is a bunch of 3D convolution layers to extract features while the decoder is a bunch of 3D deconvolution layers to generate colorful video clips from the extracted feature.

The color coherence in videos is a strong supervision signal. However, only a few work studied to employ it for self-supervised video feature learning \cite{videocolorize}. More work can be done by studying using color coherence as a supervision signal for self-supervised video feature learning.

\subsubsection{Learning from Video Prediction}

\begin{figure}[!ht]
\begin{center}
\includegraphics[width=0.5\textwidth]{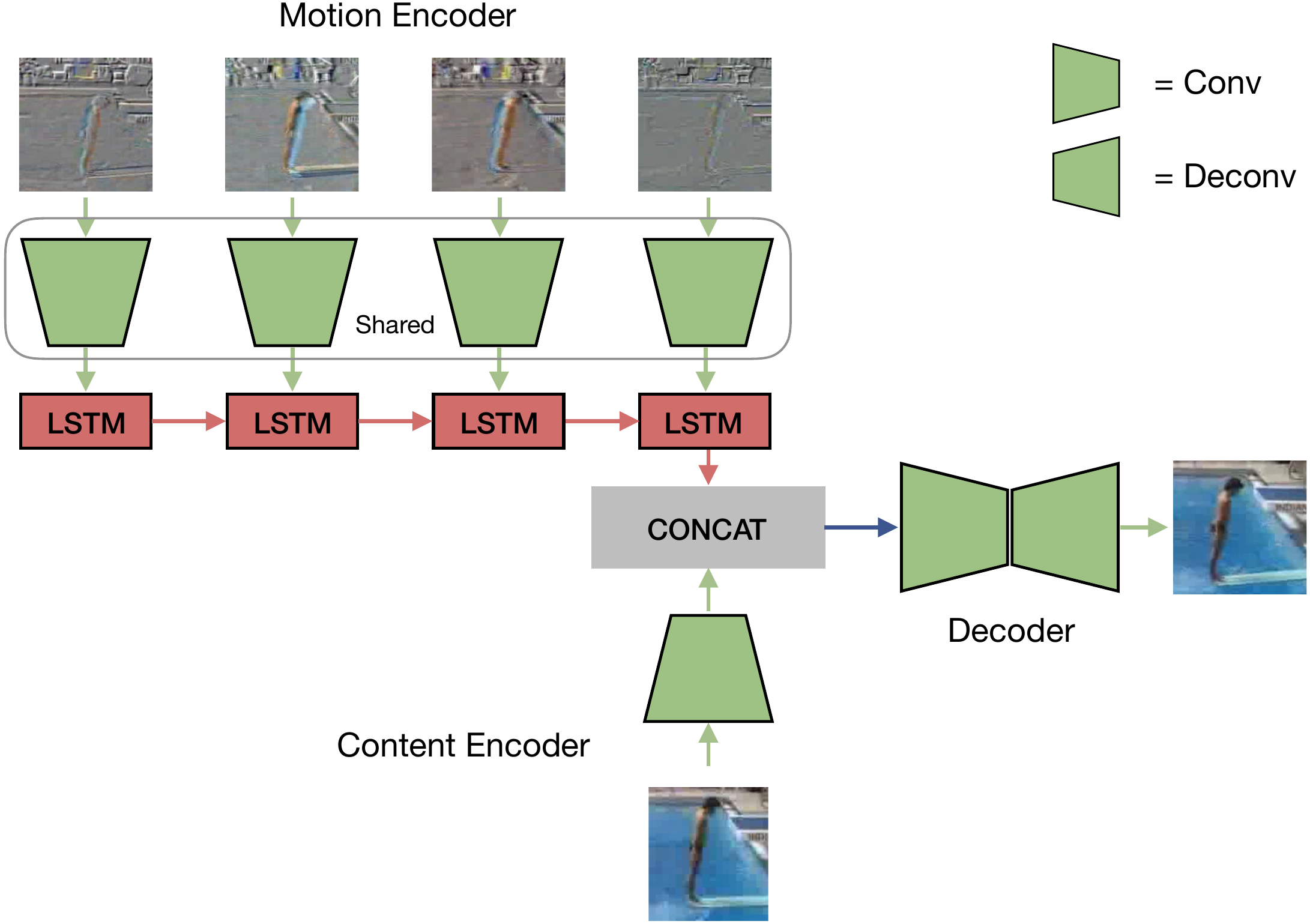}
\end{center}
\caption{The architecture for video prediction task proposed by \cite{McNet}. Figure is reproduced based on \cite{McNet}.}
\label{fig:futurepredict}
\end{figure}

Video prediction is a task of predicting future frame sequences based on a limited number of frames of a video. To predict future frames, network must learn the change in appearance within a given frame sequence. The pioneer of applying deep learning for video prediction is Un-LSTM \cite{Self-LSTM}. Due to the powerful ability of modeling long-term dynamic in videos, LSTM is used in both the encoder and decoder \cite{Self-LSTM}. 

Many methods have been proposed for video prediction \cite{Self-LSTM, McNet, MultiScaleFuture, SDC-Net, StochasticPrediction, DualMotionGAN, PhysicalInteraction}. Since its superior ability to model temporal dynamics, most of them use LSTM or LSTM variant to encode temporal dynamics in videos or to infer the future frames \cite{Self-LSTM, McNet, ConvLSTM, DualMotionGAN, PhysicalInteraction}. These methods can be employed for self-supervised feature learning without using human-annotations.

Most of the frameworks follow the encoder-decoder pipeline in which the encoder to model spatial and temporal features from the given video clips and the decoder to generate future frames based on feature extracted by encoder. Fig.~\ref{fig:futurepredict} shows a pipeline of MCnet proposed by Villegas \textit{et al.} in \cite{McNet}. McNet is built on Encoder-Decoder Convolutional Neural Network and Convolutional LSTM for video prediction. It has two encoders, one is Content Encoder to capture the spatial layout of an image, and the other is Motion Encoder to model temporal dynamics within video clips. The spatial features and temporal features are concatenated to feed to the decoder to generate the next frame. By separately modeling temporal and spatial features, this model can effectively generate future frames recursively.

Video prediction is a self-supervised learning task and the learned features can be transferred to other tasks. However, no work has been done to study the generalization ability of features learned by video prediction. Generally, The Structural Similarity Index (SSIM) and Peak Signal to Noise Ratio (PSNR) are employed to evaluate the difference between the generated frame sequence and the ground truth frame sequence. 

\subsection{Temporal Context-based Learning}

\textbf{\begin{figure}[!ht]
\begin{center}
\includegraphics[width=0.5\textwidth]{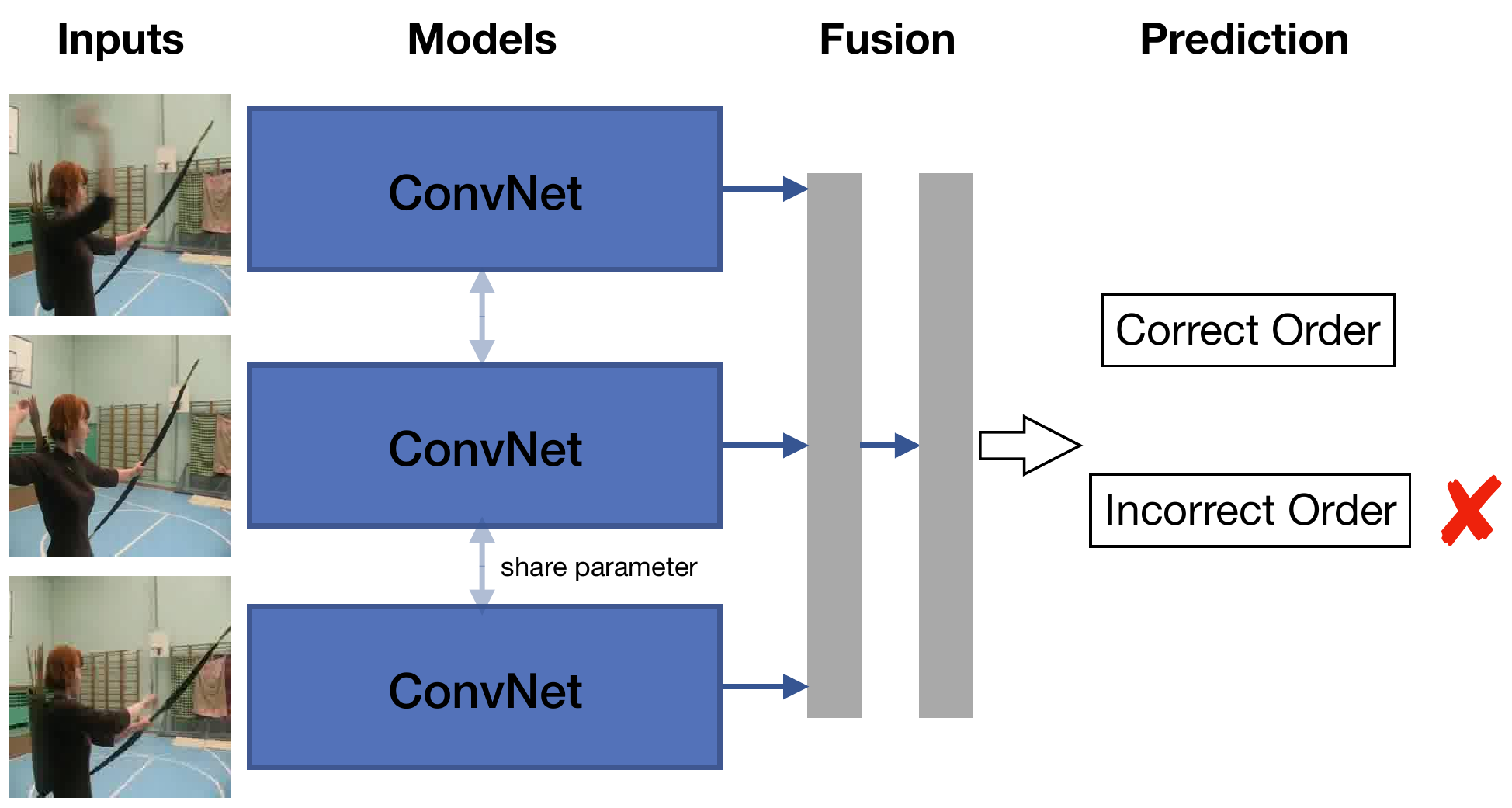}
\end{center}
\caption{The pipeline of Shuffle and Learn \cite{shuffleandlearn}. The network is trained to verify whether the input frames are in correct temporal order. Figure is reproduced based on \cite{shuffleandlearn}.}
\label{fig:shuffleandlearn}
\end{figure}}

Videos consist of various lengths of frames which have rich spatial and temporal information. The inherent temporal information within videos can be used as supervision signal for self-supervised feature learning. Various pretext tasks have been proposed by utilizing temporal context relations including temporal order verification \cite{shuffleandlearn, O3N, arrowoftime} and temporal order recognition \cite{sortsequence, CubicPuzzles}. Temporal order verification is to verify whether a sequence of input frames is in correct temporal order, while temporal order recognition is to recognize the order of a sequence of input frames.

As shown in Fig.~\ref{fig:shuffleandlearn}, Misra \textit{et al.} proposed to use the temporal order verification as the pretext task to learn image features from videos with 2DConvNet \cite{shuffleandlearn} which has two main steps: (1) The frames with significant motions are sampled from videos according to the magnitude of optical flow, (2) The sampled frames are shuffled and fed to the network which is trained to verify whether the input data is in correct order. To successfully verify the order of the input frames, the network is required to capture the subtle difference between the frames such as the movement of the person. Therefore, semantic features can be learned through the process of accomplishing this task. The temporal order recognition tasks use networks of similar architecture.

However, the methods usually suffer from a massive dataset preparation step. The frame sequences that used to train the network are selected based on the magnitude of the optical flow, and the computation process of optical flow is expensive and slow. Therefore, more straightforward and time-efficiency methods are needed for self-supervised video feature learning.

\subsection{Cross Modal-based Learning}

Cross modal-based learning methods usually learn video features from the correspondence of multiple data streams including RGB frame sequence, optical flow sequence, audio data, and camera pose.

In addition to rich temporal and spatial information in videos, optical flow sequence can be generated to specifically indicate the motion in videos, and the difference of frames can be computed with negligible time and space-time complexity to indicate the boundary of the moving objects. Similarly, audio data also provide a useful hint about the content of videos. Based on the type of data used, these methods fall into three groups: (1) methods that learn features by using the RGB and optical flow correspondence \cite{crossmodel, crosspixel}, (2) methods that learn features by utilizing the video and audio correspondence \cite{AVTS, looklistenlearn}, (3) ego-motion that learn by utilizing the correspondence between egocentric video and ego-motor sensor signals \cite{TiedEgoMotion, learn2seebymove}. Usually, the network is trained to recognize if the two kinds of input data are corresponding to each other \cite{crossmodel, AVTS}, or is trained to learn the transformation between different modalities \cite{learn2seebymove}.

\subsubsection{Learning from RGB-Flow Correspondence }

Optical flow encodes object motions between adjacent frames, while RGB frames contain appearance information. The correspondence of the two types of data can be used to learn general features \cite{crosspixel, crossmodel, FlowNet, FlowNet2}. This type of pretext tasks include optical flow estimation \cite{FlowNet, FlowNet2} and RGB and optical flow correspondence verification \cite{crosspixel}.

Sayed \textit{et al.} proposed to learn video features by verifying whether the input RGB frames and the optical flow corresponding to each other. Two networks are employed while one is for extracting features from RGB input and another is for extracting features from optical flow input \cite{crossmodel}. To verify whether two input data correspond to each other, the network needs to capture mutual information between the two modalities. The mutual information across different modalities usually has higher semantic meaning compared to information which is modality specific. Through this pretext task, the mutual information that invariant to specific modality can be captured by ConvNet.

Optical flow estimation is another type of pretext tasks that can be used for self-supervised video feature learning. Fischer \textit{et al.} proposed FlowNet which is an end-to-end convolution neural network for optical flow estimation from two consecutive frames \cite{FlowNet, FlowNet2}. To correctly estimate optical flow from two frames, the ConvNet needs to capture appearance changes of two frames. Optical flow estimation can be used for self-supervised feature learning because it can be automatically generated by simulators such as game engines or by hard-code programs without human annotation.

\subsubsection{Learning from Visual-Audio Correspondence}

Recently, some researchers proposed to use the correspondence between visual and audio streams to design “Visual-Audio Correspondence” learning task \cite{looklistenlearn, AVTS, AudioVisual, AmbientSound}.

\begin{figure}[!ht]
\begin{center}
\includegraphics[width=0.5\textwidth]{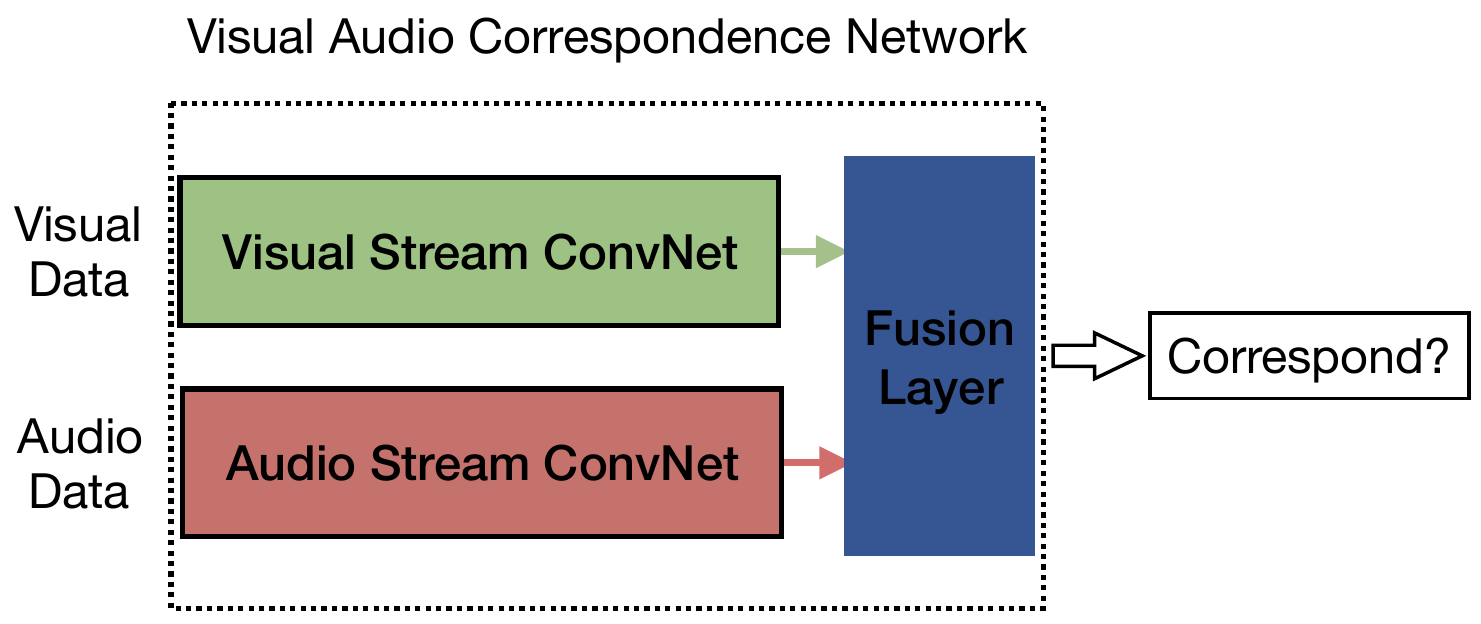}
\end{center}
\caption{The architecture of video and audio correspondence verification task \cite{looklistenlearn}.}
\label{fig:AVT}
\end{figure}

The general framework of this type of pretext tasks is shown in Fig.~\ref{fig:AVT}. There are two subnetworks: the vision subnetwork and the audio subnetwork. The input of vision subnetwork is a single frame or a stack of image frames and the vision subnetwork learns to capture visual features of the input data. The audio network is a 2DConvNet and the input is the Fast Fourier Transform (FFT) of the audio from the video. Positive data are sampled by extracting video frames and audio from the same time of one video, while negative training data are generated by extracting video frames and audio from different videos or from different times of one video. Therefore, the networks are trained to discover the correlation of video data and audio data to accomplish this task.


Since the inputs of the ConvNets are two kinds of data, the networks are able to learn the two kinds of information jointly by solving the pretext task. The performance of the two networks obtained very good performance on the downstream applications \cite{AVTS}.

\subsubsection{Ego-motion}

With the self-driving car which usually equipped with various sensors, the large-scale egocentric video along with ego-motor signal can be easily collected with very low cost by driving the car in the street. Recently, some researchers proposed to use the correspondence between visual signal and motor signal for self-supervised feature learning \cite{egomotion, learn2seebymove, TiedEgoMotion}. 

\begin{figure}[!ht]
\begin{center}
\includegraphics[width=0.5\textwidth]{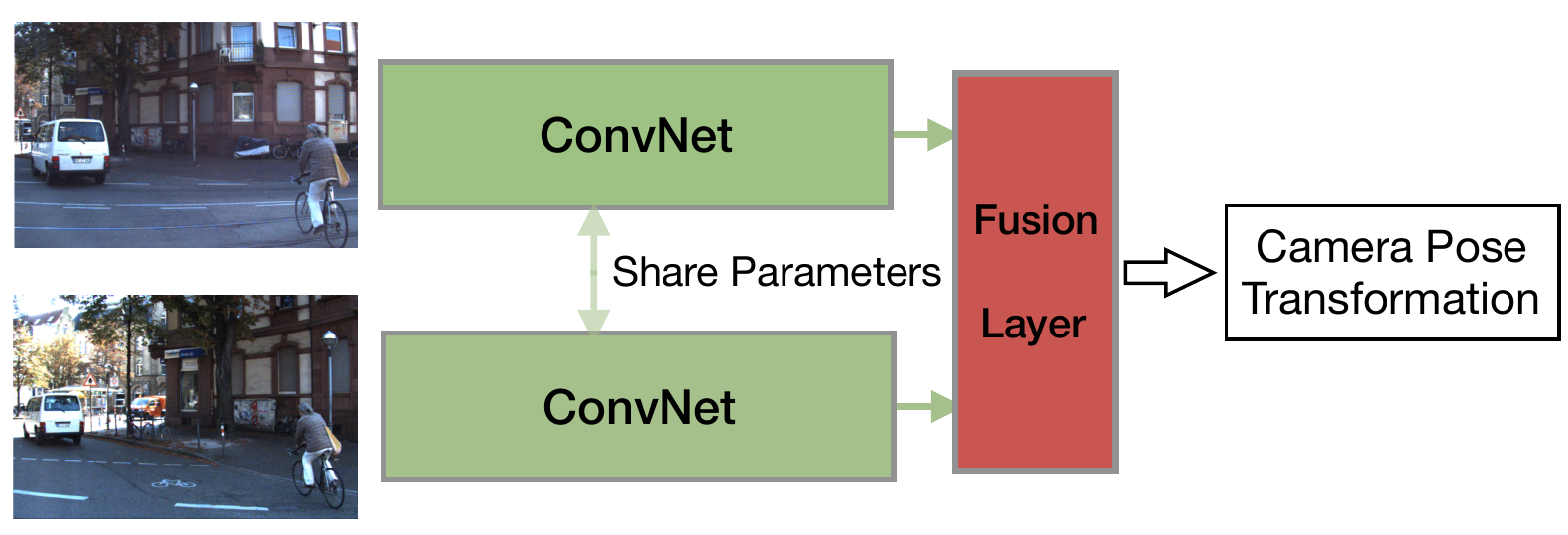}
\end{center}
\caption{The architecture of camera pose transformation estimation from egocentric videos \cite{learn2seebymove}.}
\label{fig:ego}
\end{figure}

The underline intuition of this type of methods is that a self-driving car can be treated as a camera moving in a scene and thus the egomotion of the visual data captured by the camera is as same as that of the car. Therefore, the correspondence between visual data and egomotion can be utilized for self-supervised feature learning. A typical network of using ego-motor signal is shown in Fig.~\ref{fig:ego} proposed by Agrawal \textit{et al.} for self-supervised image feature learning \cite{learn2seebymove}. The inputs to the network are two frames sampled from an egocentric video within a short time. The labels for the network indicate the rotation and translation relation between the two sampled images which can be derived from the odometry data of the dataset. With this task, the ConvNet is forced to identify visual elements that are present in both sampled images.

The ego-motor signal is a type of accurate supervision signal. In addition to directly applying it for self-supervised feature learning, it has also been used for unsupervised learning of depth and ego-motion \cite{egomotion}. All these networks can be used for self-supervised feature learning and transferred for downstream tasks.

\section{Performance Comparison}

This section compares the performance of image and video feature self-supervised learning methods on public datasets. For image feature self-supervised learning, the performance on downstream tasks including image classification, semantic segmentation, and object detection are compared. For video feature self-supervised learning, the performance on a downstream task which is human action recognition in videos is reported.

\subsection{Performance of Image Feature Learning}

As described in Section 4.3, the quality of features learned by self-supervised learned models is evaluated by fine-tuning them on downstream tasks such as semantic segmentation, object detection, and image classification. This section summarizes the performance of the existing image feature self-supervised learning methods.

Table~\ref{tab:linear} lists the performance of image classification performance on ImageNet \cite{ImageNet} and Places \cite{Places} datasets. During self-supervised pretext tasks training, most of the methods are trained on ImageNet dataset with AlexNet as based network without using the category labels. After pretext task self-supervised training finished, a linear classifier is trained on top of different frozen convolutional layers of the ConvNet on the training split of ImageNet and Places datasets. The classification performances on the two datasets are used to demonstrate the quality of the learned features.

As shown in Table~\ref{tab:linear}, the overall performance of the self-supervised models is lower than that of models trained either with ImageNet labels or with Places labels. Among all the self-supervised methods, the DeepCluster \cite{deepcluster} achieved the best performance on the two dataset. Three conclusions can be drawn based on the performance from the Table: (1) The features from different layers are always benefited from the self-supervised pretext task training. The performance of self-supervised learning methods is always better than the performance of the model trained from scratch. (2) All of the self-supervised methods perform well with the features from conv3 and conv4 layers while performing worse with the features from conv1, conv2, and conv5 layers. This is probably because shallow layers capture general low-level features, while deep layers capture pretext task-related features. (3) When there is a domain gap between dataset for pretext task training and the dataset of downstream task, the self-supervised learning method is able to reach comparable performance with the model trained with ImageNet labels.
    
\begin{table*}[ht]
  \caption{
    Linear classification on ImageNet and Places datasets using activations from the convolutional layers of an AlexNet as features. "Convn" means the linear classifier is trained based on the n-th convolution layer of AlexNet. "Places Labels" and "ImageNet Labels" indicate using supervised model trained with human-annotated labels as the pre-trained model. }
  \centering
    \begin{tabular}{@{}l c ccccc c ccccc@{}}
      \hline
            &~~~& \multicolumn{5}{c}{ImageNet} &~~~& \multicolumn{5}{c}{Places} \\
      Method &Pretext Tasks& \texttt{conv1} & \texttt{conv2} & \texttt{conv3} & \texttt{conv4} & \texttt{conv5} && \texttt{conv1} & \texttt{conv2} & \texttt{conv3} & \texttt{conv4} & \texttt{conv5} \\
      \hline
      \textbf{Places labels} \cite{AlexNet}   &--- & ---   & ---   & ---   & ---   & ---             && $22.1$ & $35.1$ & $40.2$ & $43.3$ & $44.6$  \\
      \textbf{ImageNet labels} \cite{AlexNet}  &--- & $19.3$ & $36.3$ & $44.2$ & $48.3$ & $50.5$           && $22.7$ & $34.8$ & $38.4$ & $39.4$ & $38.7$  \\
      \hline
      Random(Scratch) \cite{AlexNet}    &---  & $11.6$ & $17.1$ & $16.9$ & $16.3$ & $14.1$           && $15.7$ & $20.3$ & $19.8$ & $19.1$ & $17.5$  \\
      ColorfulColorization~\cite{colorfulcolorization} &Generation & $12.5$ & $24.5$ & $30.4$ & $31.5$ & $30.3$           && $16.0$ & $25.7$ & $29.6$ & $30.3$ & $29.7$  \\
      BiGAN~\cite{BiGAN}       &Generation & $17.7$ & $24.5$ & $31.0$ & $29.9$ & $28.0$           && $21.4$ & $26.2$ & $27.1$ & $26.1$ & $24.0$  \\
      SplitBrain~\cite{splitbrain} &Generation & $17.7$ & $29.3$ & $35.4$ & $35.2$ & $32.8$           && $21.3$ & $30.7$ & $34.0$ & $34.1$ & $32.5$  \\

      ContextEncoder~\cite{contextencoder}  &Context & $14.1$ & $20.7$ & $21.0$ & $19.8$ & $15.5$           && $18.2$ & $23.2$ & $23.4$ & $21.9$ & $18.4$  \\
      ContextPrediction~\cite{contextprediction} &Context & $16.2$ & $23.3$ & $30.2$ & $31.7$ & $29.6$           && $19.7$ & $26.7$ & $31.9$ & $32.7$ & $30.9$  \\
      Jigsaw~\cite{Jigsaw} &Context & $\textbf{18.2}$ & $28.8$ & $34.0$ & $33.9$ & $27.1$  && $23.0$ & $32.1$ & $35.5$ & $34.8$ & $31.3$  \\
      Learning2Count~\cite{Learning2Count}    &Context & $18.0$ & $30.6$ & $34.3$ & $32.5$ & $25.7$           && $\textbf{23.3}$ & $\textbf{33.9}$ & $36.3$ & $34.7$ & $29.6$  \\
      \textbf{DeepClustering}~\cite{deepcluster}  &\textbf{Context} & $13.4$ & $\textbf{32.3}$ & $\textbf{41.0}$ & $\textbf{39.6}$ & $\textbf{38.2}$ && $19.6$ & $33.2$ & $\textbf{39.2}$ & $\textbf{39.8}$ & $\textbf{34.7}$ \\
      
      \hline
    \end{tabular}
  \label{tab:linear}
\end{table*}

\begin{table*}[ht]
\small
  \centering
  \caption{
    Comparison of the self-supervised image feature learning methods on classification, detection, and segmentation on \textsc{Pascal} VOC dataset. "ImageNet Labels" indicates using supervised model trained with human-annotated labels as the pre-trained model.}
  \begin{tabular}{lc c cc c cc c cc}
  \hline
    
    Method &Pretext Tasks & Classification & & Detection & & Segmentation \\
    \hline
    \textbf{ImageNet Labels} \cite{AlexNet}    &---   & $~79.9~$ &       & $~56.8~$ && $~48.0~$ \\ 
    \hline
    Random(Scratch) \cite{AlexNet}    &---  & $~57.0~$ &  & $~44.5~$ & & $~30.1~$ \\
    
    ContextEncoder ~\cite{contextencoder} &Generation    & $~56.5~$ &  & $~44.5~$ & & $~29.7~$ \\
    BiGAN ~\cite{BiGAN}    &Generation  & $~60.1~$ &   & $~46.9~$ && $~35.2~$ \\
    ColorfulColorization ~\cite{colorfulcolorization}  &Generation  & $~65.9~$ & & $~46.9~$ && $~35.6~$ \\ 
    SplitBrain~\cite{splitbrain} &Generation & $~67.1~$ &   & $~46.7~$ && $~36.0~$ \\
    RankVideo ~\cite{wang2015unsupervised}  &Context   & $~63.1~$ &  & $~47.2~$ & & $~35.4^\dagger$   \\
    PredictNoise ~\cite{predictnoise} &Context  & $~65.3~$ & & $~49.4~$ & & $~37.1^\dagger$ \\ 
    JigsawPuzzle~\cite{Jigsaw}   &Context     & $~67.6~$ & & $~53.2~$ && $~37.6~$  \\
    ContextPrediction~\cite{contextprediction} &Context  & $~65.3~$ &  & $~51.1~$ && --- \\
    Learning2Count~\cite{Learning2Count} &Context & $~67.7~$ & & $~51.4~$ && $~36.6~$ \\
    \textbf{DeepClustering} \cite{deepcluster} &\textbf{Context}  & $\textbf{73.7}$ & & $\textbf{55.4}$ && $\textbf{45.1}$ \\
    WatchingVideo ~\cite{watchingmove}   &Free Semantic Label    & $~61.0~$ &   & $~52.2~$ & & ---     \\
    CrossDomain ~\cite{crossdomain}   &Free Semantic Label    & $~68.0~$ &   & $~52.6~$ & & ---     \\ 
    AmbientSound ~\cite{AmbientSound}   &Cross Modal & $~61.3~$ & & --- &  & ---   \\
    TiedToEgoMotion ~\cite{TiedEgoMotion}   &Cross Modal & --- & &$41.7$ &  & ---   \\
    EgoMotion ~\cite{learn2seebymove}   &Cross Modal & $~54.2~$ & & $43.9$ &  & ---   \\
    \hline
  \end{tabular}
  \label{tab:voc}
\end{table*}

In addition to image classification, object detection and semantic segmentation are also used as the downstream tasks to evaluate the quality of the features learned by self-supervised learning. Usually, ImageNet is used for self-supervised pretext task pre-training by discarding category labels, while the AlexNet is used as the base network and fine-tuned on the three tasks. Table~\ref{tab:voc} lists the performance of image classification, object detection, and semantic segmentation tasks on the PASCAL VOC dataset. The performance of classification and detection is obtained by testing the model on the test split of PASCAL VOC 2007 dataset, while the performance of semantic segmentation is obtained by testing the model on the validation split of PASCAL VOC 2012 dataset.

As shown in Table~\ref{tab:voc}, the performance of the self-supervised models on segmentation and detection dataset are very close to that of the supervised method which is trained with ImageNet labels during pre-training. Specifically, the margins of the performance differences on the object detection and semantic segmentation tasks are less than $3$\% which indicate that the learned features by self-supervised learning have a good generalization ability. Among all the self-supervised learning methods, the DeepClustering \cite{deepcluster} obtained the best performance on all the tasks.

\subsection{Performance of Video Feature Learning}

\begin{table}[!ht]
\begin{center}

\caption{Comparison of the existing self-supervised methods for action recognition on the UCF101 and HMDB51 datasets. * indicates the average accuracy over three splits. "Kinetics Labels" indicates using supervised model trained with human-annotated labels as the pre-trained model.}
\label{tab:videocompare}
\small
\begin{tabular}{|c|c|c|c|c|}
\hline
{Method}  &Pretext Task &UCF101   &HMDB51 \\
\hline\hline
Kinetics Labels* \cite{3DResNet}   & ---   &$84.4$ &$56.4$\\
\hline
VideoGAN \cite{VideoGAN}  &Generation &$52.1$ &---\\
VideoRank \cite{wang2015unsupervised}   &Context   &$40.7$ &$15.6$\\
ShuffleLearn \cite{shuffleandlearn} &Context  &$50.9$ &$19.8$\\
OPN  \cite{O3N}  &Context &$56.3$ &$22.1$ \\
RL \cite{RL} &Context  &$58.6$ &$25.0$ \\
AOT \cite{arrowoftime} &Context  &$58.6$ &--- \\
3DRotNet \cite{3DRotNet} &Context &$62.9$ &$\textbf{33.7}$\\
\textbf{CubicPuzzle*} \cite{CubicPuzzles} &\textbf{Context} &$\textbf{65.8}$ &$\textbf{33.7}$\\
RGB-Flow \cite{crossmodel} &Cross Modal &$59.3$ &$27.7$\\
{PoseAction} \cite{poseaction} &Cross Modal &$55.4$ &$23.6$\\
\hline
\end{tabular} 
\end{center}
\end{table}

For self-supervised video feature learning methods, human action recognition task is used to evaluate the quality of learned features. Various video datasets have been used for self-supervised pre-training, and different network architectures have been used as the base network. Usually after the pretext task pre-training finished, networks are fine-tuned and tested on the commonly used UCF101 and HMDB51 datasets for human action recognition task. Table ~\ref{tab:videocompare} compares the performance of existing self-supervised video feature learning methods on UCF101 and HMDB51 datasets.

As shown in Table~\ref{tab:videocompare}, the best performance of the fine-tune results on UCF101 is less than $66$\%. However, the supervised model which trained with Kinetics labels can easily obtain an accuracy of more than $84$\%. The performance of the self-supervised model is still much lower than the performance of the supervised model. More effective self-supervised video feature learning methods are desired. 

\subsection{Summary}

Based on the results, conclusions can be drawn about the performance and reproducibility of the self-supervised learning methods.

\textbf{Performance:} For image feature self-supervised learning, due to the well-designed pretext tasks, the performance of self-supervised methods are comparable to the supervised methods on some downstream tasks, especially for the object detection and semantic segmentation tasks. The margins of the performance differences on the object detection and semantic segmentation tasks are less than $3$\% which indicate that the learned features by self-supervised learning have a good generalization ability. However, the performance of video feature self-supervised learning methods is still much lower than that of the supervised models on downstream tasks. The best performance of the 3DConvNet-based methods on UCF101 dataset is more than $18$\% lower than that of the supervised model \cite{3DResNet}. The poor performance of 3DCovnNet self-supervised learning methods probably because 3DConvNets usually have more parameters which lead to easily over-fitting and the complexity of video feature learning due to the temporal dimension of the video.

\textbf{Reproducibility:} As we can observe, for the image feature self-supervised learning methods, most of the networks use AlexNet as a base network to pre-train on ImageNet dataset and then evaluate on same downstream tasks for quality evaluation. Also, the code of most methods are released which is a great help for reproducing results. However, for the video self-supervised learning, various datasets and networks have been used for self-supervised pre-training, therefore, it is unfair to directly compare different methods. Furthermore, some methods use UCF101 as self-supervised pre-training dataset which is a relatively small video dataset. With this size of the dataset, the power of a more powerful model such as 3DCovnNet may not be fully discovered and may suffer from server over-fitting. Therefore, larger datasets for video feature self-supervised  pre-training should be used. 

\textbf{Evaluation Metrics:} Another fact is that more evaluation metrics are needed to evaluate the quality of the learned features in different levels. The current solution is to use the performance on downstream tasks to indicate the quality of the features. However, this evaluation metric does not give insight what the network learned through the self-supervsied pre-training. More evaluation metrics such as network dissection \cite{NetDissection} should be employed to analysis the interpretability of the self-supervised learned features. 

\section{Future Directions}

Self-supervised learning methods have been achieving great success and obtaining good performance that close to supervised models on some computer vision tasks. Here, some future directions of self-supervised learning are discussed.

\textbf{Learning Features from Synthetic Data: } A rising trend of self-supervised learning is to train networks with synthetic data which can be easily rendered by game engines with very limited human involvement. With the help of game engines, millions of synthetic images and videos with accuracy pixel-level annotations can be easily generated. With accurate and detailed annotations, various pretext tasks can be designed to learn features from synthetic data. One problem needed to solve is how to bridge the domain gap between synthetic data and real-world data. Only a few work explored self-supervised learning from synthetic data by using GAN to bridge the domain gap \cite{crossdomain, LearnFromGames}. With more available large-scale synthetic data, more self-supervised learning methods will be proposed.

\textbf{Learning from Web Data:} Another rising trend is to train networks with web collected data \cite{Youtube8M, WebVision, VisualText} based on their existing associated tags. With the search engine, millions of images and videos can be downloaded from websites like Flickr and YouTube with negligible cost. In addition to its raw data, the title, keywords, and reviews can also be available as part of the data which can be used as extra information to train networks. With carefully curated queries, the web data retrieved by reliable search engines can be relatively clean. With large-scale web data and their associated metadata, the performance of self-supervised methods may be boosted up. One open problem about learning from web data is how to handle the noise in web data and their associated metadata. 

\textbf{Learning Spatiotemporal Features from Videos:} Self-supervised image feature learning has been well studied and the margin of the performance between supervised models and self-supervised models are very small on some downstream tasks such as semantic segmentation and object detection. However, self-supervised video spatiotemporal feature learning with 3DConvNet is not well addressed yet. More effective pretext tasks that specifically designed to learn spatiotemporal features from videos are needed.

\textbf{Learning with Data from Different Sensors:} Most existing self-supervised visual feature learning methods focused on only images or videos. However, if other types of data from different sensors are available, the constraint between different types of data can be used as additional sources to train networks to learn features \cite{egomotion}. The self-driving cars usually are equipped with various sensors including RGB cameras, gray-scale cameras, 3D laser scanners, and high-precision GPS measurements and IMU accelerations. Very large-scale datasets can be easily obtained through the driving, and the correspondence of data captured by different devices can be used as a supervision signal for self-supervised feature learning.

\textbf{Learning with Multiple Pretext Tasks:} Most existing self-supervised visual feature learning methods learn features by training ConvNet to solve one pretext tasks. Different pretext tasks provide different supervision signals which can help the network learn more representative features. Only a few work explored the multiple pretext tasks learning for self-supervised feature learning \cite{multitasklearning, crossdomain}. More work can be done by studying the multiple pretext task self-supervised feature learning.

\section{Conclusion}

Self-supervised image feature learning with deep convolution neural network has obtained great success and the margin between the performance of self-supervised methods and that of supervised methods on some downstream tasks becomes very small. This paper has extensively reviewed recently deep convolution neural network-based methods for self-supervised image and video feature learning from all perspectives including common network architectures, pretext tasks, algorithms, datasets, performance comparison, discussions, and future directions etc. The comparative summary of the methods, datasets, and performance in tabular forms clearly demonstrate their properties which will benefit researchers in the computer vision community.

\bibliographystyle{ieeetr}
\bibliography{SelfSupervised}
\end{document}